\pdfoutput=1
\documentclass[11pt]{article}

\usepackage[final]{acl}

\makeatletter
\newcommand{\anonymize}[2]{\ifacl@anonymize #2\else #1\fi}
\makeatother

\newcommand{\finalsection}[1]{\anonymize{#1}{}}

\usepackage{times}
\usepackage{latexsym}

\usepackage[T1]{fontenc}
\usepackage[utf8]{inputenc}

\usepackage{microtype}

\usepackage{inconsolata}

\usepackage{graphicx}
\usepackage{amssymb}

\usepackage{xparse}
\usepackage{tabularx}
\usepackage{soul}
\usepackage{amsmath}
\usepackage{booktabs}
\usepackage[most]{tcolorbox}
\tcbuselibrary{breakable,listings,skins}
\usepackage{cuted}
\usepackage{caption}
\usepackage{placeins}
\usepackage{enumitem}
\usepackage{dblfloatfix}
\usepackage{multirow}
\usepackage{float}
\usepackage{algorithm}
\usepackage{algpseudocode} % from algorithmicx
\usepackage{booktabs}
\usepackage{makecell}

\title{\longtitle}

\author{
Muhammad Umair \and JP de Ruiter\\ 
Department of Computer Science\\
Tufts University\\  
Medford, Massachusetts, USA\\
\texttt{\{muhammad.umair, jp.deruiter\}@tufts.edu}
}

\usepackage{xspace}

\newcommand{\longtitle}{Using Semantic Uncertainty to Estimate Transition Relevance in Turn-taking\xspace}

\newcommand{\hide}[1]{}

\newcommand{\word}[1]{w_{#1}}

\newcommand{\turnLen}{N}

\newcommand{\prefix}[1]{P_{#1}}
\newcommand{\prefixes}{\mathcal{P}_\stimulus}

\newcommand{\TRP}[1]{T_{#1}}

\newcommand{\TRPhat}[1]{\widehat{T}_{#1}}

\newcommand{\TRPs}{\mathbf{T}_{\stimulus}}

\newcommand{\TRPsHat}{\widehat{\mathbf{T}}_{\stimulus}}
\newcommand{\stimulus}{S}
\newcommand{\response}{R}

\newcommand{\interval}[2]{I_{#1,#2}}

\newcommand{\intervalProportion}[2]{\interval{#1}{#2}^{Proportion}}

\newcommand{\uncertainty}{U}

\newcommand{\uncertaintyAt}[1]{\uncertainty_{#1}}

\newcommand{\uncertaintyVector}{\mathbf{\uncertainty}_{\stimulus}}

\newcommand{\uncertaintyGradient}{\Delta \uncertaintyVector}

\newcommand{\trpfunc}{f}

\newcommand{\uncertaintyPrefix}[1]{\left(\uncertaintyVector\right)_{1:#1}}

\newcommand{\trpfuncEq}[1]{\TRPhat{#1} = \trpfunc\!\left(\uncertaintyPrefix{#1}\right)}

\newcommand{\continuationsLen}{K}
\newcommand{\prefixContinuations}[1]{A_{#1}}
\newcommand{\singleContinuation}[2]{a_{#1,#2}}

\newcommand{\singleSentenceEmbed}[2]{h_{#1,#2}}

\newcommand{\prefixSimMatrix}[1]{S^{(#1)}}

\newcommand{\uncertaintyChangeWindow}[1]{W_{#1}}

\newcommand{\smoothedUncertaintyVector}{\tilde{\uncertaintyVector}}

\newcommand{\smoothedUncertaintyGradient}{\Delta \smoothedUncertaintyVector}

\newcommand{\smoothedUncertaintyGradAt}[1]{\left(\smoothedUncertaintyGradient\right)_{#1}}

\newcommand{\smoothingOperator}{\mathcal{F}}

\begin{document}

\maketitle

\begin{abstract}
  Turn-taking is a fundamental mechanism that governs when interlocutors speak and listen. Although Spoken Dialogue Systems (SDS) exploit a range of linguistic, acoustic, and non-verbal cues, they produce ill-timed responses in unscripted interaction. A central challenge is anticipating Transition Relevance Places (TRPs), or \emph{opportunities}, not obligations, for a listener to take the floor. Human listeners do not wait for turn endings; as an utterance unfolds, they use \emph{expectations} about its developing meaning to anticipate TRPs and decide whether to take the floor. We examine whether these evolving expectations can be modeled through \emph{semantic uncertainty}—an LLM-derived measure of how strongly a turn so far constrains what may plausibly come next. To do so, we sample possible continuations of an ongoing turn and use changes in semantic dispersion to identify TRPs \emph{within} turns. We evaluate this account on a dataset with TRP labels derived from real-time listener responses, rather than retrospective annotation. Our approach substantially outperforms prompt-based and fine-tuned text-only baselines, providing empirical support for the view that evolving semantic constraints inform perceived turn-taking opportunities in unscripted interaction.

\end{abstract}

\section{Introduction}
\label{sec:intro}

% Highlight specific properties of turn-taking that enable smooth interaction. 
Humans coordinate speaking and listening through a sophisticated turn-taking mechanism \citep{sacks1974simplest}. Unlike in formal settings, where who speaks when may be predetermined, speaker selection in unscripted interaction is managed on a per-turn basis. This depends on conversationalists’ orientation to \emph{Transition Relevance Places} (TRPs): points at which a turn may be treated as possibly complete and a listener may, but is not obligated to, respond \citep{Levinson1983Pragmatics,selting2000construction}.

A key distinction concerns whether a TRP results in a speaker change. If a listener takes the floor, the TRP occurs \emph{between} turns and is directly observable. Many TRPs, however, arise \emph{within} turns, where a listener could have responded but did not \citep{selting2000construction}. These within-turn TRPs are an integral part of the turn-taking system but leave limited empirical trace in recorded data \citep{umair-etal-2024-large,castillo2025survey}.

% - Why is the TRP issue important for dialogue systems? 
This creates a challenge for Spoken Dialogue Systems (SDS), which continue to produce ill-timed responses and stilted feedback in unscripted interaction \citep{algherairy2024review,patamia2025turn,arora2025talking}. Models trained on observable turn-taking behavior (speaker changes, backchannels, etc.) receive direct supervision for TRPs that listeners acted on, but not for the broader set of response opportunities they may have perceived \citep{umair-etal-2024-large}. This motivates a representational question: what information do humans use to anticipate TRPs, and can that be used to support their prediction in dialogue systems?

Human listeners do not wait for a turn to end; they recognize in advance where it may be complete. This prospective recognition is known as \emph{projection} \citep{sacks1974simplest,schegloff1996turn}. As each new word arrives, what can be projected changes, and listeners revise their \emph{expectations} about how the turn may continue \citep{magyari2012prediction,riest2015anticipation}.

We operationalize these evolving expectations as \emph{semantic uncertainty}: an LLM-derived measure of how an unfolding turn constrains what may plausibly come next \citep{shorinwa2025survey}. We estimate this signal by sampling possible continuations of the turn so far and measuring their semantic dispersion \citep{nguyen-etal-2025-beyond}. We ask whether changes in this uncertainty predict TRPs more accurately than direct baselines. Our contribution is an empirical study of whether explicitly tracking evolving semantic constraint, derived from text alone, supports the prediction of perceived response opportunities in unscripted interaction.

% What are previous works / features of this work that are directly relevant? 
\section{Related Work}
\label{sec:related_work}

\subsection{Turn-Taking and the Role of Semantics}
\label{subsec:semantics_in_turn_taking}

% What are TRPs?
Transition Relevance Places (TRPs) are points in an utterance at which a listener could, but is not obligated to, initiate a response \citep{sacks1974simplest}. At these locations, a listener may take the floor, the current speaker may continue, or a listener may produce minimal contributions, such as backchannels (e.g., hmm, uh-huh; \citeauthor{yngve1970backchannel} \citeyear{yngve1970backchannel}) or continuers (e.g., yeah, okay; \citeauthor{schegloff1982discourse} \citeyear{schegloff1982discourse}). Depending on its timing, such a response may encourage the current speaker to continue or perform a specific conversational action \citep{schegloff1982discourse}. Overlapping talk is likewise not necessarily interruptive. Whether an entry is heard as competitive depends partly on where it begins relative to the developing turn and on how participants manage the overlap \citep{schegloff2000overlapping,drew2009}. While turn-timing varies across cultures, the basic organization of rapid speaker transition is broadly shared across languages \citep{stivers2009universals}. Given the importance of TRPs for coordinating turns, a central question emerges: how do listeners project TRPs as an utterance unfolds?

% Go into why lexico-syntactic / semantic information.
Experimental work suggests that, alongside prosodic and nonverbal cues, a turn's developing lexico-syntactic structure is particularly important for projecting possible completion \citep{deRuiter2006ProjectingTheEnd,levinson2016turn}. Listeners accurately projected turn endings when lexico-syntactic content was preserved but pitch was flattened to a monotone, whereas performance declined when words were made unrecognizable but intonation was retained \citep{deRuiter2006ProjectingTheEnd}. They also anticipate upcoming words in a turn, the accuracy of which is closely linked to how well they estimate when a turn will end \citep{magyari2012prediction}.

% They also anticipate words within a turn, the accuracy of which is closely linked to how well they estimate when a turn will end \citep{magyari2012prediction}.

% What is the role of smenaitc expectations? 
Psycholinguistic accounts describe projection as an incremental process in which listeners form and revise expectations about how a turn may continue \citep{riest2015anticipation,levinson2016turn,magyari2012prediction}. Each new word changes the lexico-syntactic and semantic continuations that listeners may expect \citep{tanenhaus1995integration,altmann1999incremental,hale2001probabilistic,levy2008expectation}. Some words narrow these possibilities, while extensions, repairs, qualifications, and redirections may introduce new ones and change how listeners expect the turn to develop \citep{lerner1991syntax,schegloff1996turn,selting2000construction,liddicoat2004projectability}. 

Changes in semantic uncertainty may therefore reflect how the range of plausible meanings evolves over the course of a turn. A decrease indicates that plausible continuations are becoming more similar in meaning, whereas an increase indicates greater variation. Such changes may be relevant to within-turn TRPs because semantic convergence may support anticipation of a continuation, while divergence may signal that an earlier expectation needs to be revised or deferred \citep{magyari2012prediction,riest2015anticipation}. 

% What about other cues? 
Multimodal cues, such as prosody, gaze, and gesture, also play an important \emph{disambiguating} role in TRP projection. These cues are not interpreted independently of the unfolding semantics and pragmatics of a turn \citep{kendrick2023turn}. Rather, they help listeners distinguish possible completion from continuation and assess whether a response is relevant at a particular position \citep{holler2019multimodal}. They are particularly informative in complex situations such as overlap, interruption, and miscommunication \citep{skantze2014turn,Torreira2015IntonationalBoundaries}.

\subsection{Computational Models of Turn-Taking}
\label{subsec:models_of_turn_taking}

% What are the current computational models used for TRP prediction?
Computational work has largely modeled turn-taking through observable interactional outcomes. TurnGPT predicts token-level turn-shift probabilities from text and speaker identity, while RC-TurnGPT additionally conditions on a candidate system response \citep{ekstedt2020turngpt,jiang-etal-2023-response}. Because both are trained on speaker changes, their targets are realized between-turn transitions rather than within-turn TRPs, which need not result in a listener taking the floor \citep{threlkeld2022using}.

% What about VAP-based models?
Voice Activity Projection (VAP) models take a complementary approach by forecasting joint future speech activity from acoustic signals. Turn holds, speaker switches, overlaps, and mutual silence can be derived from these forecasts \citep{ekstedt2022much}. VAP has been extended to multimodal, multi-party, and multilingual settings and combined with text-based approaches \citep{onishi2024multimodal,inoue2024multilingual,leishman2024pairwiseturngpt,wang2024turn,elmers2025triadic}.

Despite these advances, dialogue systems struggle to time responses \citep{algherairy2024review,patamia2025turn,arora2025talking}. Predicting voice activity or speaker changes may therefore not fully capture when a response becomes interactionally relevant \citep{liesenfeld-etal-2023-timing}.

% What is the modeling issue?
This gap is difficult to study because many richly annotated interaction corpora organize turns around speaker changes \citep{anderson1991hcrc,jurafsky1997switchboard,calhoun2010nxt,reece2023candor}, while task-oriented corpora use system--user alternation \citep{budzianowski2018multiwoz,si2023spokenwoz}. These resources therefore primarily capture where speakers \emph{did} respond rather than where they \emph{could have} responded but did not.

\subsection{Semantic Uncertainty Quantification}
\label{subsec:SUQuantification}

Black-box uncertainty quantification estimates variation in model outputs without requiring access to model output distributions, which may not be available for state-of-the-art LLMs \citep{liesenfeld2024rethinking,shorinwa2025survey}. Conventional measures such as normalized predictive entropy can conflate variation in meaning with variation in lexical form by treating semantically equivalent paraphrases as distinct outcomes \citep{malinin2020uncertainty,kuhn2023semantic}. Cluster-based semantic uncertainty methods address this by grouping meaning-equivalent outputs into semantic classes via entailment and computing entropy over those classes \citep{kuhn2023semantic,farquhar2024detecting}. More recent variants replace hard clustering with graded similarity through kernel-based functions \citep{nikitin2024kernel} or pairwise sentence-embedding similarities \citep{nguyen-etal-2025-beyond}.

These approaches have primarily been applied in tasks such as question answering, summarization, and translation \citep{kuhn2023semantic,farquhar2024detecting,shorinwa2025survey}. Less is known about whether these measures can characterize how an utterance becomes more or less semantically constrained as it unfolds.

\section{Approach}
\label{sec:approach}

\subsection{Within-Turn TRP Prediction Task}
\label{subsec:within_turn_trp_task}

% ---- UPDATED 

% Q: What is the input object that the task is operating on? 
Following \citet{umair-etal-2024-large}, we define our task as predicting \emph{opportunities}, not obligations, for response within a single speaker's unfolding turn. As an utterance unfolds, the task is to predict whether it affords a possible listener response, regardless of whether a listener actually takes the floor.

Formally, we define a single speaker's turn as a \emph{stimulus} $\stimulus = \langle \word{1}, \ldots, \word{\turnLen} \rangle$, a sequence of $\turnLen$ words, where the number of words may vary across stimuli. We further segment each stimulus into a series of \emph{prefixes}, where each prefix $\prefix{i} = \langle \word{1}, \ldots, \word{i} \rangle$ consists of the sequence of words from the first word through $\word{i}$. We denote by $\prefixes = \langle \prefix{1}, \ldots, \prefix{\turnLen} \rangle$ the ordered set of all prefixes derived from a stimulus $\stimulus$.

For each prefix $\prefix{i}$, we associate a binary \emph{reference label} $\TRP{i} \in \{0,1\}$ indicating whether the position immediately following its last word $\word{i}$ is labeled as a TRP. The resulting sequence $\TRPs = \langle \TRP{1}, \ldots, \TRP{\turnLen} \rangle$ constitutes a reference TRP labeling for a stimulus, with $\TRP{\turnLen}$ corresponding to the turn-final position. TRP predictions are therefore conditioned only on preceding linguistic material, reflecting the \emph{causal} constraints under which listeners form TRP judgments. Section~\ref{subsec:data} describes the procedure by which the reference labels $\TRPs$ are constructed from the listener-response dataset.

%----- ALGORITHM 1, associated with 'Estimating Semantic Uncertainty' section. Here so it appears on top of the page with that section. 

\begin{algorithm*}[tb]
\caption{Estimating Semantic Uncertainty for a Prefix}
\label{alg:semantic_uncertainty}
\begin{algorithmic}[1]
\Require
Prefix $\prefix{i}=\langle \word{1},\ldots,\word{i}\rangle$;
language model $\mathcal{M}$;
number of continuations $\continuationsLen\in\mathbb{N}_{\ge1}$;
maximum continuation length $L\in\mathbb{N}_{\ge1}$;
embedding model $\mathcal{E}:\text{Seq}\rightarrow\mathbb{R}^d$;
similarity scaling parameter $\tau>0$
\Ensure
Semantic uncertainty value $\uncertaintyAt{i} \in \mathbb{R}$

\State Sample $\continuationsLen$ stochastic continuations
$\prefixContinuations{i} = \{\singleContinuation{i}{1}, \ldots, \singleContinuation{i}{\continuationsLen}\}$
from $\mathcal{M}$ conditioned on $\prefix{i}$, each with maximum length $L$ words

\For{$j = 1$ to $\continuationsLen$}
    \State Compute continuation embedding
    $\singleSentenceEmbed{i}{j} \gets \mathcal{E}(\singleContinuation{i}{j})$
\EndFor

\State Construct semantic similarity matrix
$\prefixSimMatrix{i} \in \mathbb{R}^{\continuationsLen \times \continuationsLen}$,
where
$\prefixSimMatrix{i}_{u,v}
= \cos\!\left(\singleSentenceEmbed{i}{u}, \singleSentenceEmbed{i}{v}\right)$

\State Compute semantic uncertainty
$\uncertaintyAt{i} \gets \mathrm{SNNE}(\prefixSimMatrix{i})$
\Comment{see Eq.~(\ref{eq:snne})}

\State \Return $\uncertaintyAt{i}$
\end{algorithmic}
\end{algorithm*}

%----- 

\newtheorem{definition}{Definition}[section]

\begin{definition}[Within-turn TRP Prediction]
\label{def:within_turn_trp_task}
Given a stimulus $\stimulus$ and its associated prefix sequence $\prefixes$, produce a predicted binary label sequence $\TRPsHat = \langle \TRPhat{1}, \ldots, \TRPhat{\turnLen} \rangle$, where each $\TRPhat{i} \in \{0,1\}$ indicates whether the position immediately following $\word{i}$ is predicted to afford a possible transition, based solely on the prefix $\prefix{i}$. Predictions are evaluated against the corresponding reference labeling $\TRPs$.
\end{definition}

% NOTE: Placing here so it appears at the right location. 

\subsection{Predicting TRPs via Semantic Uncertainty}
\label{subsec:projecting_trps_via_semantic_uncertainty}
% --- UPDATED ---
\vspace{-0.8mm}
% Q: What is semantic uncertainty? Why is it relevant? 

To support TRP prediction, we introduce \emph{semantic uncertainty} as an intermediate representation. For each prefix $\prefix{i}$, semantic uncertainty is a scalar value $\uncertaintyAt{i} \in \mathbb{R}$ that reflects how strongly the prefix constrains what may plausibly come next. We use \emph{semantic} to refer to variation in meaning among possible model outputs, rather than variation in surface form alone \citep{kuhn2023semantic,farquhar2024detecting,nguyen-etal-2025-beyond}. Intuitively, $\uncertaintyAt{i}$ is low when the possible continuations of $\prefix{i}$ are similar in meaning, and high when they differ more substantially. Applied across the prefix sequence $\prefixes$, this yields a causal trajectory $\uncertaintyVector = \langle \uncertaintyAt{1},\ldots,\uncertaintyAt{\turnLen}\rangle$ that tracks how the space of plausible continuations narrows or widens as the turn unfolds. Section~\ref{subsec:estimating_semantic_uncertainty} describes how we estimate this quantity.

% Q: How do we use these uncertainty values? As sequences? 
We do not associate TRPs with the absolute uncertainty value of a prefix. Instead, we relate TRPs to \emph{changes} in uncertainty as the turn unfolds (Section~\ref{subsec:semantics_in_turn_taking}). These changes reflect shifts in how strongly the utterance constrains its plausible continuations, highlighting locations where a listener may, but is not obligated to, respond. 

% Q: Once we have obtained the sequence of uncertainty values, how do we wactually use them? 
We further define a TRP decision function $\trpfunc$ that maps a causal, variable-length prefix of semantic uncertainty values to a binary label. For $i = 1, \ldots, \turnLen$, this label is given by $\trpfuncEq{i}$, where $\uncertaintyPrefix{i} = \langle \uncertaintyAt{1}, \ldots, \uncertaintyAt{i} \rangle$. This formulation is agnostic to the specific choice of $\trpfunc$: it may operate over absolute uncertainty values, local changes, or aggregated statistics over time. Section~\ref{subsec:decision_rule} describes our concrete instantiation.

\subsection{Estimating Semantic Uncertainty}
\label{subsec:estimating_semantic_uncertainty}

We estimate $\uncertaintyAt{i}$ using the procedure summarized in Algorithm~\ref{alg:semantic_uncertainty}. For each prefix $\prefix{i}$, we sample continuations from a language model and compare their similarity in a sentence-embedding space. We then aggregate their pairwise similarities using \emph{Semantic Nearest Neighbor Entropy} \citep[SNNE; see Equation~\ref{eq:snne};][]{nguyen-etal-2025-beyond}. SNNE is appropriate here because it uses sampled continuations rather than token-level probability distributions, and because it compares continuations using pairwise semantic similarity instead of first grouping them into entailment-based semantic clusters (\citealt{kuhn2023semantic,farquhar2024detecting,nguyen-etal-2025-beyond}; see Section~\ref{subsec:SUQuantification}). The resulting scalar $\uncertaintyAt{i}$ reflects the semantic dispersion of the continuations. We treat this embedding-space dispersion as a model-mediated proxy for semantic variation, without assuming independence from lexical form. Appendix~\ref{appsubsec:embedding_validity} provides support for this treatment. 

\begin{align}
\mathrm{SNNE}(\prefixSimMatrix{i})
&=
-\frac{1}{\continuationsLen}
\sum_{u=1}^{\continuationsLen}
\log
\left(
\sum_{v=1}^{\continuationsLen}
\exp\!\left(
\frac{\prefixSimMatrix{i}_{u,v}}{\tau}
\right)
\right)
\label{eq:snne}
\end{align}

We measure semantic proximity $\prefixSimMatrix{i}_{u,v}$ using cosine similarity, a standard measure for comparing sentence embeddings \citep{reimers2019sentence}. The similarity scaling parameter $\tau$ controls how strongly SNNE is influenced by nearest neighbors in the similarity matrix: smaller values emphasize the most similar continuation pairs, while larger values distribute weight more broadly across pairwise similarities. Under this formulation, more negative SNNE values indicate more semantically constrained continuations, while less negative values indicate greater indeterminacy. Applying Algorithm~\ref{alg:semantic_uncertainty} across all prefixes of a stimulus yields a prefix-indexed uncertainty trajectory $\uncertaintyVector = \langle \uncertaintyAt{1}, \ldots, \uncertaintyAt{\turnLen} \rangle$, which tracks how semantic constraints evolve as the speaker’s turn unfolds.

\begin{algorithm*}[t]
\caption{Causal TRP decision rule for a Prefix $\TRPhat{i}=\trpfunc(\uncertaintyPrefix{i})$}
\label{alg:trp_decision}
\begin{algorithmic}[1]
\Require
Index $i \in \{1,\ldots,N\}$;
Uncertainty prefix $\uncertaintyPrefix{i}=\langle \uncertaintyAt{1},\ldots,\uncertaintyAt{i}\rangle \in \mathbb{R}^i$;
Window size $w \in \mathbb{N}_{\ge 1}$;
Causal smoothing operator $\smoothingOperator$;
Threshold $\theta \in \mathbb{R}_{>0}$ ;
Constant $\varepsilon>0$
\Ensure
Predicted label $\TRPhat{i}\in \{0,1\}$

\State $j \gets \max(2, i-w+1)$
\If{$j \ge i$} \Return $0$ \EndIf

\State $\uncertaintyGradient \gets \langle 0,\ \uncertaintyAt{2}-\uncertaintyAt{1},\ldots,\uncertaintyAt{i}-\uncertaintyAt{i-1}\rangle$
\State $\smoothedUncertaintyGradient \gets \smoothingOperator(\uncertaintyGradient)$
\Comment{$\smoothingOperator$ uses only indices $\le i$}

\State $\uncertaintyChangeWindow{i} \gets \langle \smoothedUncertaintyGradAt{j},\ldots,\smoothedUncertaintyGradAt{i-1}\rangle$
\State $b \gets \mathrm{median}(\uncertaintyChangeWindow{i})$,\quad
$s \gets \mathrm{MAD}(\uncertaintyChangeWindow{i})$
\Comment{$\mathrm{MAD}(X)=\mathrm{median}(|X-\mathrm{median}(X)|)$}

\State $z \gets \dfrac{|\smoothedUncertaintyGradAt{i}-b|}{s+\varepsilon}$
\State \Return $\TRPhat{i} \gets \mathbb{I}[z>\theta]$
\end{algorithmic}
\end{algorithm*}

\subsection{Rule over Uncertainty Dynamics}
\label{subsec:decision_rule}

% --- UPDATED

% Q: What is the decision function and what does it capture?
We define our decision function $\trpfunc$ based on
the view that TRPs arise from local shifts in semantic
constraint (see Section~\ref{subsec:semantics_in_turn_taking}).
Accordingly, $\trpfunc$ identifies candidate TRPs from both convergent
shifts, where the utterance becomes more constrained, and divergent
shifts, where unexpected extensions occur. We adopt a
deterministic rule to emphasize interpretability and leave more expressive realizations, such as learned models, for future work.

% Q: How is this rule operationalized?
Algorithm~\ref{alg:trp_decision} operationalizes this intuition. It defines uncertainty change as discrete differences between successive prefixes. Because uncertainty estimates are based on finite continuation samples, the change signal may contain sampling noise rather than genuine shifts. The algorithm therefore attenuates this noise while preserving local change structure using a low-pass filter. It evaluates the current smoothed change relative to a trailing window of past changes to obtain a locally adaptive baseline. It quantifies how exceptional the current change is using the median and Median Absolute Deviation (MAD) and predicts a TRP when the normalized value exceeds a fixed threshold.

\section{Experimental Setup}
\label{sec:experimental_setup}

\subsection{Data: Empirical Within-Turn TRPs}
\label{subsec:data}

% Q: What data are you using, and why is it appropriate?
We use the participant-response dataset\footnote{Stimulus and participant-response audios are available at \url{https://osf.io/k5pc9/overview?view_only=5124d862448f4435b775d49a7b299d6d}.} released by \citet{umair-etal-2024-large}, which consists of 55 single-speaker stimuli and responses from 118 participants. In the original study, the stimuli were organized into four presentation lists: two original lists and a reversed-order version of each. The reversed lists were used to counterbalance stimulus-order effects, which refer to differences in response behavior caused by where a stimulus appeared in the sequence \citep{deRuiter2006ProjectingTheEnd,riest2015anticipation}.

The elicitation paradigm captured \emph{within-turn} response opportunities through real-time listener responses rather than post-hoc annotations or observed speaker changes. Each participant was randomly assigned to one list and asked to verbalize brief backchannels (e.g., `hmm', `yes') when they perceived an opportunity for speech.

\begin{figure*}[tb]
    \centering
    \includegraphics[width=1\linewidth]{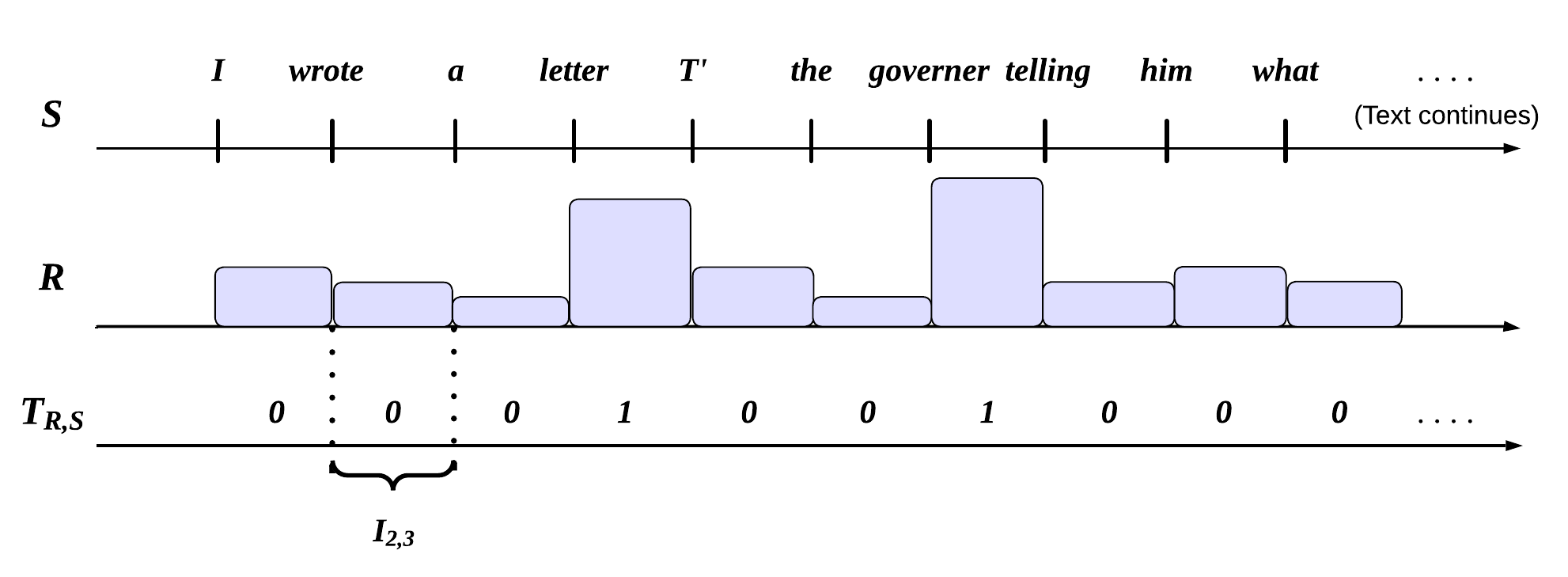}
    \vspace{-2em}
    \caption{
    Schematic illustration of empirical within-turn TRP labeling.
    $\stimulus$ denotes the speaker's word sequence, $\response$ denotes aggregated listener response distribution, and $\TRPs$ denotes the resulting binary reference labels. Participant response onsets are aligned to word-adjacent intervals $\interval{i}{i+1}$, each corresponding to the position following prefix $\prefix{i}$. Intervals are labeled positive when their empirical response proportion $\intervalProportion{i}{i+1}$ exceeds the stimulus-specific threshold. The example text is adapted from \citet{schegloff1982discourse}; the figure is inspired by \citet{umair-etal-2024-large}.
    }
    \label{fig:labeling_example}
\end{figure*}

\begin{table}[tb]
\centering
\small
\setlength{\tabcolsep}{6pt}
\renewcommand{\arraystretch}{1.1}
\begin{tabular}{@{}lr@{}}
\toprule
\textbf{Statistic} & \textbf{Value} \\
\midrule
\multicolumn{2}{@{}l}{\textit{Released dataset \citep{umair-etal-2024-large}}} \\
\quad Stimuli & 55 \\
\quad Participants & 118 \\
\quad Participants per stimulus, median (range) & 58 (51--60) \\
\addlinespace
\multicolumn{2}{@{}l}{\textit{Reconstructed positions and labels}} \\
\quad Candidate TRP positions (total prefixes) & 5{,}195 \\
\quad TRP-positive positions & 842 \\
\quad Positive-label prevalence (842 / 5{,}195) & 16.2\% \\
\addlinespace
\multicolumn{2}{@{}l}{\textit{Diagnostic}} \\
\quad Local maxima among positive labels (of 842) & 72\% \\
\bottomrule
\end{tabular}
\caption{
Dataset statistics. The 16.2\% prevalence is computed over all candidate positions; the 72\% local-maximum statistic is computed over the 842 positive labels only and is diagnostic rather than definitional.
}
\label{tab:dataset_stats}
\end{table}

We derive the reference TRP labels $\TRPs$ by aligning participant response onsets with the speaker's word-level timing. \citet{umair-etal-2024-large} released stimulus and per-participant response recordings, but not the orthographic transcripts, word-level stimulus timings, or response-onset timings required here. We reconstructed these manually using ELAN and Praat \citep{wittenburg-etal-2006-elan,Boersma2009}. We excluded clear non-response vocalizations, such as breaths, laughter, and throat clearings. Because the original study reported no order effects, we pooled responses across original and reversed presentations of each stimulus, yielding a mean of 59 participants per stimulus (median 58; see Table~\ref{tab:dataset_stats}).

For each stimulus, we define word-adjacent intervals $\langle \interval{1}{2}, \ldots, \interval{\turnLen-1}{\turnLen}\rangle$. Each interval $\interval{i}{i+1}$ corresponds to the candidate TRP position immediately following prefix $\prefix{i}$ and its reference label $\TRP{i}$. We compute its empirical response proportion $\intervalProportion{i}{i+1}$ as the fraction of participants who responded within the interval, counting each participant at most once (see Figure~\ref{fig:labeling_example}).

To account for temporally dispersed response onsets and varying baseline responsiveness, we binarize these proportions using a stimulus-specific threshold. Specifically, $\TRP{i}=1$ when $\intervalProportion{i}{i+1} > \mu + 0.05\sigma$, where $\mu$ and $\sigma$ are the mean and standard deviation of the nonzero response proportions for that stimulus; otherwise, $\TRP{i}=0$. Zero-response intervals are excluded because they would pull $\mu$ toward zero and make the threshold uninformative. The turn-final position $\TRP{\turnLen}$ is always labeled positive. This yields 842 positive positions out of 5{,}195 (16.2\%; see Table~\ref{tab:dataset_stats}). Among these, 72\% are local maxima relative to adjacent intervals. This statistic is diagnostic, not definitional: local maximality is not part of the labeling procedure. Additional diagnostics are provided in Appendix~\ref{appsec:trp_label_construction}; implications of the dataset size are discussed in Section~\ref{sec:limitations}.

\subsection{Evaluation Metrics}
\label{subsec:eval_metrics}

The sparse nature of intervals in our data labeled as TRPs (16.2\% of 5{,}195; see Table~\ref{tab:dataset_stats}) highlights that TRP prediction is highly imbalanced. When predictions control system turn entry, a false positive may cause the system to interrupt the user's ongoing turn, whereas a missed TRP may delay a response \citep{skantze2021turnreview,arora2025talking}. For dialogue systems, false-positive TRP predictions may therefore carry greater cost than missed ones.

Accuracy is uninformative in this setting; majority-class predictors can score well while failing to detect rare but meaningful events \citep{he2009learning}. Instead, we report $F_{0.5}$, which weighs precision more heavily than recall, and the True Negative Rate (TNR), which measures suppression of spurious predictions. For comparability with prior work, we report balanced accuracy, which treats false positives and negatives symmetrically and therefore does not reflect the asymmetric interactional cost of mistimed turn entries. Section~\ref{sec:limitations} discusses the limitations of this approach.

% Q: How do we apply this in our case? 
Across experiments, we report aggregated metrics using a two-level macro-average rather than pooling predictions across all stimuli. Because stimuli vary in length and number of TRPs, pooling predictions would allow the most TRP-dense stimuli to dominate. We therefore compute each metric independently for each stimulus, average across stimuli within each list, and then average across lists. This gives equal weight to each stimulus and each list in the final aggregate. 

\section{Experiments and Results}
\label{sec:experiments}

\subsection{Prompt-Based TRP Prediction}
\label{subsec:exp_prompt_baselines}

\begin{table*}[t]
\centering
\small
\renewcommand{\arraystretch}{1.2}
\setlength{\tabcolsep}{3.5pt}
\begin{tabular}{lcccccccccccccccc}
\toprule
& \multicolumn{4}{c}{\textbf{Expert}}
& \multicolumn{4}{c}{\textbf{Participant}}
& \multicolumn{4}{c}{\textbf{Imagined}}
& \multicolumn{4}{c}{\textbf{Oracle}} \\
\cmidrule(lr){2-5}
\cmidrule(lr){6-9}
\cmidrule(lr){10-13}
\cmidrule(lr){14-17}
\textbf{Model}
& $F_{0.5}$ & P & TNR & BA
& $F_{0.5}$ & P & TNR & BA
& $F_{0.5}$ & P & TNR & BA
& $F_{0.5}$ & P & TNR & BA \\
\midrule
LLaMA-3.1-8B
& 0.12 & 0.12 & 0.81 & 0.5
& 0.10 & 0.12 & 0.84 & 0.49
& 0.18 & 0.16 & 0.48 & 0.50
& 0.17 & 0.16 & 0.62 & 0.49 \\
LLaMA-3.1-70B
& \textbf{0.18} & 0.22 & 0.86 & 0.53
& \textbf{0.22} & 0.21 & 0.79 & 0.54
% & 0.18 & 0.16 & 0.48 & 0.50
& \textbf{0.21} & 0.24 & 0.90 & 0.54
& 0.14 & 0.17 & 0.90 & 0.53 \\
% & 0.17 & 0.16 & 0.62 & 0.49 \\
Mistral-7B
& 0.10 & 0.10 & 0.79 & 0.50
& 0.12 & 0.12 & 0.77 & 0.51
& 0.20 & 0.17 & 0.35 & 0.53
& \textbf{0.19} & 0.16 & 0.01 & 0.49 \\
Mixtral-8x7B
& 0.06 & 0.13 & 0.97 & 0.50
& 0.04 & 0.11 & 0.99 & 0.50
& 0.10 & 0.16 & 0.91 & 0.51
& 0.17 & 0.15 & 0.53 & 0.51 \\
Qwen2.5-7B
& 0.09 & 0.11 & 0.82 & 0.50
& 0.05 & 0.09 & 0.92 & 0.50
& 0.11 & 0.15 & 0.91 & 0.51
& 0.01 & 0.02 & 0.97 & 0.50 \\
Qwen2.5-72B
& 0.08 & 0.20 & 0.99 & 0.51
& 0.08 & 0.14 & 0.97 & 0.51
& 0.21 & 0.31 & 0.93 & 0.54
& 0.16 & 0.22 & 0.88 & 0.52 \\
\bottomrule
\end{tabular}
\caption{
Prompt-based within-turn TRP prediction across instruction-tuned models (see Table~\ref{tab:models_evaluated}) and prompting conditions. Expert and participant are direct prompting conditions; imagined and oracle are future-context variants of the participant-style prompt. P denotes precision and BA denotes balanced accuracy. Metrics are macro-averaged as described in Section~\ref{subsec:eval_metrics}. The highest $F_{0.5}$ value within each prompting condition is bolded.
}
\label{tab:prompting_results_all_models_main}
\end{table*}

We first test whether stronger instruction-tuned models improve prompt-based TRP prediction relative to \citet{umair-etal-2024-large}. Given mixed evidence on whether detailed task-specific prompts improve performance (see \citealt{reynolds2021prompt,webson2022prompt,xu2023expertprompting}), we compare four prompting conditions. The \emph{expert} condition provides an explicit, theory-driven definition of TRPs, while the \emph{participant} condition mirrors the listener instructions used to collect the dataset. Inspired by response-conditioned models (see \citealt{jiang-etal-2023-response}), we also test two future-context variants: an \emph{imagined} condition, where the model generates a plausible continuation before predicting, and an \emph{oracle} condition, where the true upcoming same-speaker words are provided.

Table~\ref{tab:prompting_results_all_models_main} reports results for all six instruction-tuned models (see Appendix~\ref{appsec:models_and_infra}) across the four prompting conditions. Across models, performance remains weak: balanced accuracy stays near chance, and both precision and $F_{0.5}$ are low. Although the imagined condition yields the best within-model $F_{0.5}$ for four of six models, no prompting condition reliably identifies TRPs. Oracle access to the true upcoming words also does not consistently improve performance. Overall, direct prompting does not recover within-turn TRPs, even when possible or actual same-speaker future context is available. Experimental details are in Appendix~\ref{appsec:prompt_trp_full_results}; prompt templates are in Appendix~\ref{app:prompts}. The prompts include no demonstrations from the evaluation dataset, making the setup zero-shot with respect to the target distribution \citep{brown2020language}.

\subsection{Supervised Fine-Tuning (SFT) Based TRP Prediction}
\label{subsec:exp_SFT}

We next use supervised fine-tuning (SFT), which, unlike prompting, updates model weights, to test whether explicit task adaptation improves within-turn TRP prediction. Because prompting shows no consistent benefit from model scale, we fine-tune LLaMA-3.1-8B-Instruct with LoRA adapters rather than the strongest prompt-based model, LLaMA-3.1-70B-Instruct (Table~\ref{tab:prompting_results_all_models_main}; \citealt{hu2022lora}). Larger models may behave differently under SFT; we leave this comparison to future work.

We construct training examples by pairing each prefix $\prefix{i}$ with its binary reference label $\TRP{i}$. To avoid leakage across prefixes from the same turn, we split data by stimulus rather than prefix. The train, validation, and test splits cover 60\%, 20\%, and 20\% of the 55 stimuli, yielding 3{,}098, 1{,}203, and 894 prefixes, with 505, 188, and 149 TRP-positive labels, respectively. The held-out test split preserves the original class imbalance.

To assess whether performance is limited by sparse TRP-positive training examples, we construct three nested supervision regimes---Low, Medium, and Full---from the training split. For each regime, we select a target number of TRP-positive prefixes and include adjacent negatives to preserve local context around TRP-labeled positions \citep{he2009learning}. Validation and test splits are fixed across regimes i.e., the ablation varies only the training subset (see Table~\ref{tab:sft_regime_results}).

% Q: What did we find? 
Table~\ref{tab:sft_regime_results} shows that supervision yields modest gains at the Medium regime, but improvements are limited and non-monotonic. Because results are based on a single stimulus-level split, we interpret these trends qualitatively. The pattern suggests that direct supervision provides some benefit, but performance does not consistently scale with additional data. Full training details are in Appendix~\ref{appsec:SFT_details}.

\vspace{-1mm}

\begin{table}[h]
\centering
\small
\renewcommand{\arraystretch}{1.15}
\setlength{\tabcolsep}{3.5pt}
\begin{tabular}{@{}lccc cccc@{}}
\toprule
& \multicolumn{3}{c}{\textbf{Training data}}
& \multicolumn{4}{c}{\textbf{Test performance}} \\
\cmidrule(lr){2-4}
\cmidrule(lr){5-8}
\textbf{Regime}
& \textbf{Size}
& \textbf{TRPs}
& \textbf{TRP \%}
& $\mathbf{F_{0.5}}$
& \textbf{P}
& \textbf{TNR}
& \textbf{BA} \\
\midrule
Low
& 528
& 152
& 28.79
& 0.14
& 0.17
& \textbf{0.93}
& 0.51 \\
Medium
& 1,011
& 303
& 29.97
& \textbf{0.27}
& \textbf{0.28}
& 0.89
& \textbf{0.57} \\
Full
& 1,526
& 505
& 33.09
& 0.15
& 0.21
& 0.91
& 0.50 \\
\bottomrule
\end{tabular}
\caption{
SFT training-regime TRP prediction performance for LLaMA-3.1-8B-Instruct on a held-out test set. P denotes precision and BA denotes balanced accuracy. Metrics are macro-averaged (Section~\ref{subsec:eval_metrics}); the highest value in each performance column is bolded.
}
\label{tab:sft_regime_results}
\end{table}

\subsection{Semantic-Uncertainty-Based TRP Prediction}
\label{subsec:exp_SU}

\begin{figure}[h]
    \centering
    \includegraphics[width=\columnwidth]{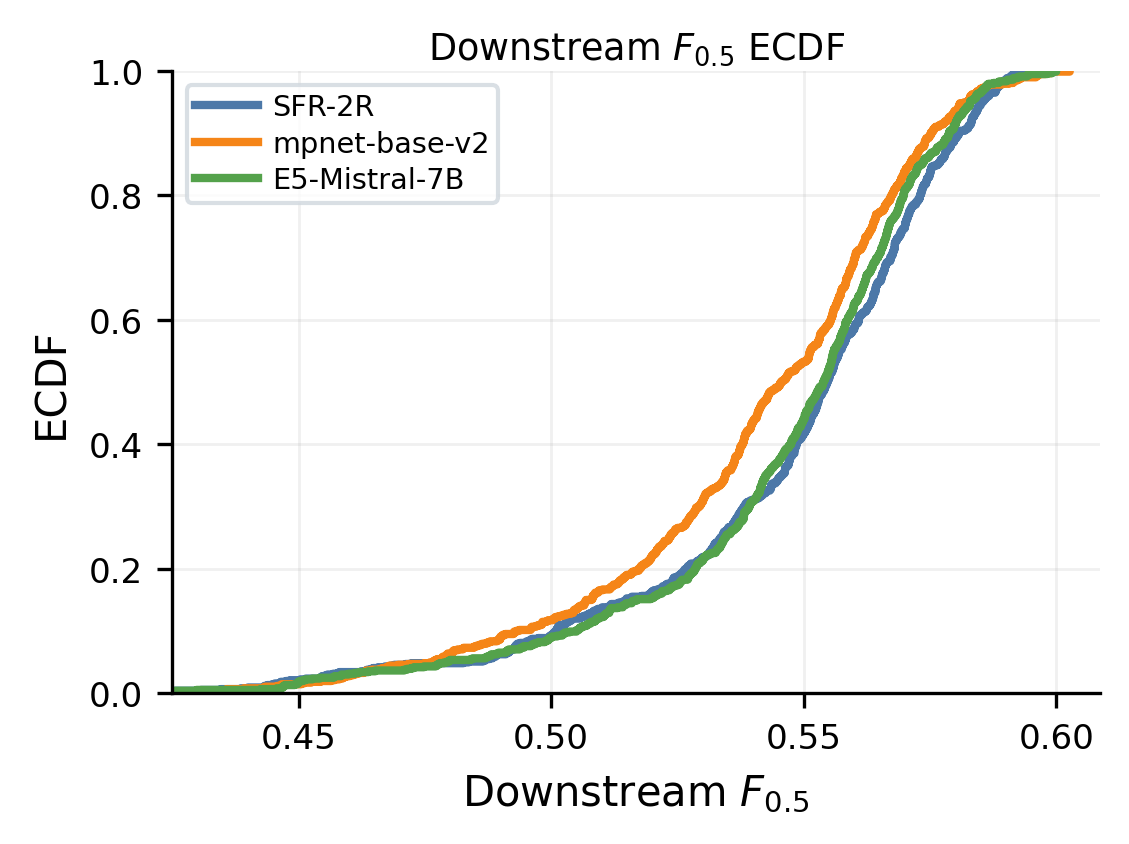}
        \caption{
Distribution of $F_{0.5}$ across the fixed-$K$ semantic-uncertainty sweep. Each curve is an empirical cumulative distribution over configurations for one embedding model. Mean $\pm$ SD across configurations: SFR-2R $= 0.547 \pm 0.033$, all-mpnet-base-v2 $= 0.540 \pm 0.033$, E5-Mistral-7B-Instruct $= 0.546 \pm 0.031$.
}
\label{fig:semantic_uncertainty_ecdf}
\end{figure}

% ------
% Q: What exactly is being tested here, and what is the experimental philosophy?

We now evaluate our proposed method (see Section~\ref{subsec:projecting_trps_via_semantic_uncertainty}). For the main evaluation, we fix sampling at $\continuationsLen = 25$ continuations per prefix as well as the sampling model (LLaMA-3.1-8B-Instruct), and sweep the remaining configurations across qualitatively distinct settings within our compute budget. The uncertainty-estimation stage (Algorithm~\ref{alg:semantic_uncertainty}) spans 6 decoding settings, 3 embedding models (110M--7B parameters), and 4 SNNE $\tau$ values, yielding 72 uncertainty-signal configurations. The decision stage (Algorithm~\ref{alg:trp_decision}) applies 32 detector configurations to each trajectory, for $72 \times 32 = 2{,}304$ total configurations (see Table~\ref{tab:semantic_uncertainty_sweep}). Grid density reflects computational cost: we limit expensive continuation sampling to 6 settings but sweep the inexpensive decision stage over saved trajectories. The grid was fixed before evaluation, and no configuration was selected using test performance; we report results across the full grid. Full configuration details appear in Appendix~\ref{subsec:SU_exp_setup_details}.

Figure~\ref{fig:semantic_uncertainty_ecdf} summarizes the main fixed-$\continuationsLen$ sweep. Across configurations, the semantic-uncertainty pipeline achieves a mean $F_{0.5}$ of 0.545 ($\mathrm{SD}=0.032$), ranging from 0.385 to 0.603. The embedding-specific distributions are closely aligned, with mean $F_{0.5}$ values ranging from 0.540 to 0.547. Appendix~\ref{appsubsec:embedding_validity} complements this comparison by showing that the embedding space captures semantic similarity across lexical variation. Results are also stable across continuation-sampling regimes: more stochastic decoding increases continuation diversity and average SNNE, while the same prefixes remain in similar semantic regions across regimes (see Appendix~\ref{subsec:SU_robustness}).

% Q: How stable is the unecertainty signal - how much information does it provide? 
A one-way $\eta^2$ decomposition shows that 75.4\% of $F_{0.5}$ variance is associated with the decision rule $\trpfunc$; smoothing choices alone explain 52.6\%. Upstream construction choices contribute much less: decoding setting, embedding model, and $\tau$ explain 5.5\%, 0.9\%, and 0.88\%, respectively. Although these one-way effects are not additive, they indicate that SNNE preserves the TRP-relevant signal across construction choices. Varying $\tau$ changes the range of the uncertainty trajectory, but has only a small downstream effect (see Appendix~\ref{subsec:tau_analysis}).

% Q: How do we know that the semantic uncertainty is the informative signal based on the variance decompsotion. 
The dominance of the decision rule and near-identical performance across embedding models raise the possibility that the gains come from the rule (Algorithm~\ref{alg:trp_decision}) rather than the semantic information carried by SNNE (Algorithm~\ref{alg:semantic_uncertainty}). To test this, we replace the SNNE trajectory with two predictive-entropy controls while holding the prefixes, decision rule, evaluation, and generation model (LLaMA-3.1-8B-Instruct) fixed. \emph{Next-token entropy} (NTE) is the Shannon entropy of the model's next-token distribution at each prefix and tests whether the rule can exploit token-level uncertainty without sampled continuations or embeddings. \emph{Normalized predictive entropy} (NPE) averages length-normalized mean token surprisal over the same $\continuationsLen=25$ continuations, testing whether their semantic dispersion adds information beyond their token probabilities. NTE and NPE achieve mean \(F_{0.5}\) scores of $0.319$ and $0.328$, respectively (see Table~\ref{tab:summary_comparison}). Their maxima, $0.356$ and $0.376$, remain below the minimum semantic-uncertainty ($F_{0.5}=0.385$), showing that the decision rule is not signal-agnostic and that the embedding-based semantic dispersion captured by SNNE contributes information beyond token-probability uncertainty.

\begin{table}[t]
\centering
\small
\renewcommand{\arraystretch}{1.15}
\setlength{\tabcolsep}{5pt}
\begin{tabularx}{\columnwidth}{@{}>{\raggedright\arraybackslash}Xc@{}}
\toprule
\textbf{Approach} & $\mathbf{F}_{0.5}$ \\
\midrule
GPT-4 Omni, participant \citep{umair-etal-2024-large}$^\dagger$
& 0.15 \\
Prompt-based prediction (best)
& 0.22 \\
Supervised fine-tuning (best)$^\ddagger$
& 0.27 \\
\addlinespace
Next-token entropy (NTE)
& $0.319 \pm 0.070$ \\
Normalized predictive entropy (NPE)
& $0.328 \pm 0.027$ \\
Semantic uncertainty
& $\mathbf{0.545 \pm 0.032}$ \\
\bottomrule
\end{tabularx}
\caption{
Summary $F_{0.5}$ comparison. Baseline rows report the best result; uncertainty-signal rows report mean $\pm$ SD across configurations (32 per alternative signal and 2{,}304 for semantic uncertainty).
$^\dagger$Approximated from the reported precision and recall under global-threshold labels.
$^\ddagger$Evaluated on the held-out split.
}
\label{tab:summary_comparison}
\end{table}

% Q: Does the result depend on the number of sampled continuations?
We next test whether performance using SNNE depends on continuation-sample size $\continuationsLen$. Larger samples may improve dispersion estimates but increase sampling cost. Holding the decision-rule sweep fixed, we vary $\continuationsLen$ starting at 2, the smallest nontrivial pairwise setting. Mean $F_{0.5}$ reaches approximately 0.54 at $\continuationsLen=2$ and remains near that level for larger samples, with a shallow optimum around four to five continuations (see Figure~\ref{fig:K_ablation}). This stability suggests that local uncertainty shifts are recoverable from small samples. Moreover, SNNE at $\continuationsLen=2$ outperforms NPE at $\continuationsLen=25$, indicating that its advantage is not explained by sample size alone. 

\begin{figure}[h]
    \centering
    \includegraphics[width=\columnwidth]{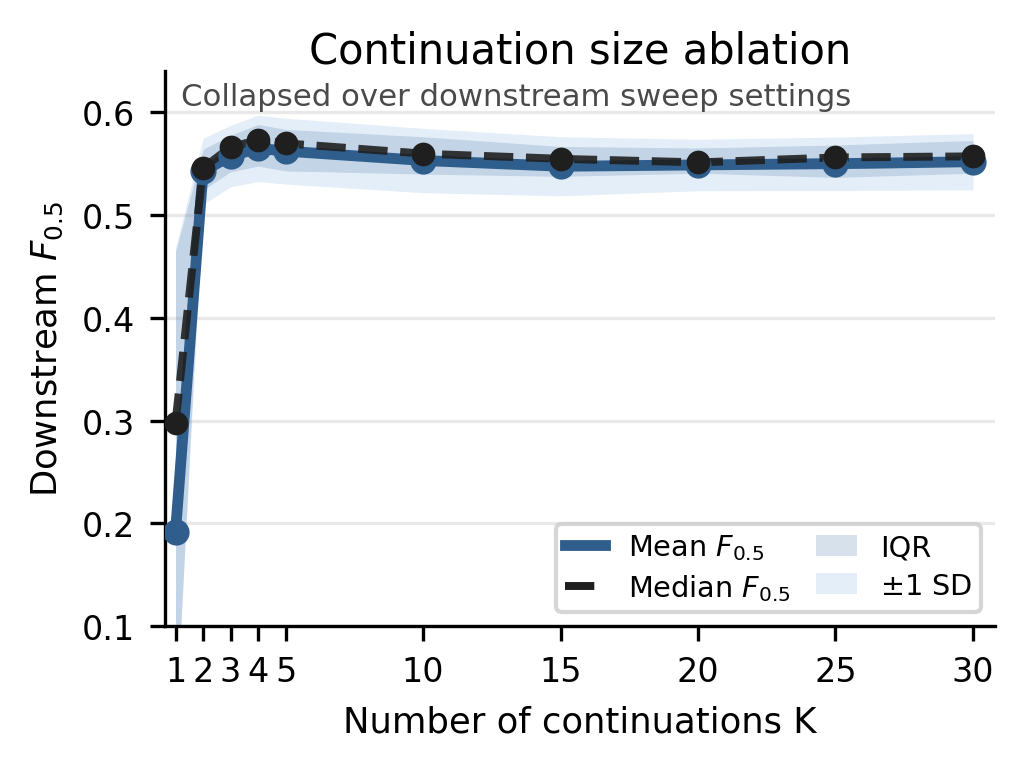}
    \caption{
    Continuation-sample size ablation. For each $K$, downstream $F_{0.5}$ is collapsed over the decision-rule sweep. Lines show mean and median performance; shaded bands show interquartile range and $\pm 1$ std.
    }
    \label{fig:K_ablation}
\end{figure}

\section{Discussion} 

% Q: What is a brief recap?
Listeners in unscripted interaction project TRPs before a speaker's turn has ended, using the turn-so-far to form expectations about what may plausibly come next. This remains difficult for dialogue systems, partly because they are typically trained on observable outcomes rather than response opportunities that do not result in speaker change.

We therefore ask whether evolving semantic constraints help identify within-turn TRPs. We estimate semantic uncertainty over sampled continuations and use local changes in its trajectory as the prediction signal. Because this procedure is computationally costly, we leave its deployment in dialogue systems as an avenue for future work.

Empirically, prompt-based inference and SFT do not reliably predict within-turn TRPs. Prompting yields low $F_{0.5}$ across models and conditions, while SFT produces only limited, non-monotonic gains. In contrast, SNNE-based prediction performs substantially better. Most performance variation is associated with the decision rule, but neither predictive-entropy control reaches the SNNE performance range under the same rule, indicating that SNNE captures TRP-relevant uncertainty beyond token-probability uncertainty. Performance remains stable from two sampled continuations onward, suggesting that informative local changes in SNNE are recoverable from small samples.

Together, these findings suggest that evolving semantic structure provides a useful signal for within-turn TRP prediction. This is notable because our labels derive from real-time listener responses rather than retrospective annotation, targeting opportunities that listeners perceived but did not necessarily act on. Semantic uncertainty should nonetheless be read as a model-mediated proxy for semantic variation, not a measurement of listener-internal expectations. For dialogue systems, this suggests that text-only TRP prediction may benefit from intermediate representations of semantic constraint.

\section{Conclusion}
\label{sec:conclusion}

% NOTE: Your conclusion should not just be a summary of what you said in the paper. Imagine the reader only reads the abstract, intro and conclusion. In this case, the abstract gets the reader interested enough to even read the paper -- i.e., the "interesting" result should be mentioned in there. The intro outlines the problem and gives a sneak peak at the contribution. The conclusion has the thing X that the reader would've learned had they read the entire paper. It might have things in there that you wouldn't understand if you didn't skim the rest of the paper. Basically, you need to reward those readers who made it through your entire paper! :)  

Even though spoken dialogue systems draw on linguistic, acoustic, and multimodal cues for turn-taking, they continue to struggle to time responses appropriately in natural interaction. One reason is that TRPs, or opportunities for speech, are not confined to turn endings. Many occur within turns and leave no interactional trace when a listener does not respond, limiting their visibility in dialogue corpora. Models trained on observable outcomes therefore receive evidence about where listeners did respond, but not necessarily about where they could have responded. Our analyses show that this mismatch has measurable consequences. Direct text-to-label prediction remains unreliable, even across diverse prompting conditions and with supervised fine-tuning. By contrast, semantic-uncertainty-based prediction provides a more reliable basis for identifying within-turn TRPs by tracking how the space of plausible continuations becomes more or less semantically dispersed as an utterance unfolds. The improvement is robust across the tested settings, but the broader contribution is representational. Semantic uncertainty shows that changes in the space of plausible continuations carry information about when response opportunities arise. Progress on turn-taking may therefore require models that track how semantic constraints evolve over a turn, not only models trained on observable response outcomes.

\section{Limitations}
\label{sec:limitations}

% What empirical claims does this dataset license, and which broader claims about turn-taking does it explicitly not support?
We acknowledge several limitations. First, our empirical evaluation is based on a small, English-only dataset of single-speaker turns labeled through participant judgments of perceived TRPs (see Section~\ref{subsec:data}). This dataset is well suited to studying within-turn TRPs, which are rarely observable in standard corpora. At the same time, it limits generalization. Our findings have not yet been validated on widely used dialogue datasets such as Switchboard or SpokenWOZ \citep{jurafsky1997switchboard,si2023spokenwoz}. This reflects a broader methodological challenge, since within-turn TRPs are underrepresented in corpora derived from natural interaction \citep{threlkeld2022using,umair-etal-2024-large}. Further validation across datasets and interactional settings is therefore required.

% Which components of human turn-taking are intentionally excluded by modeling TRP projection from text alone, and how should the results be interpreted given this exclusion?
Second, our evaluation isolates text-derived semantic information and therefore does not capture the full multimodal structure of turn-taking. This choice allows us to test whether evolving semantic constraint contributes to within-turn TRP prediction, but it does not show that semantic uncertainty alone is sufficient for response timing in deployed dialogue systems. Acoustic, prosodic, and multimodal cues may interact with semantic uncertainty in ways that are not captured here. Future work should therefore test whether the signal remains useful when integrated with models that incorporate these additional sources of information.

% To what extent is the proposed semantic uncertainty signal a property of the unfolding utterance versus a consequence of specific model, sampling, and embedding choices?
Third, even within the text-only setting, semantic uncertainty remains model-mediated. We do not claim that embedding-space dispersion is independent of lexical form, only that it provides a proxy for semantic relatedness less tied to surface form than token-level measures. A diagnostic analysis supports this claim. Duplicate-continuation rates vary sharply across decoding settings, from roughly 67\% to 1\%, while prefix-level mean embeddings remain stable, with pairwise cosine similarities of 0.94–0.97. This reduces, but does not eliminate, the concern that the signal reflects surface form. It also does not make the signal model-independent; alternative language models, embedding spaces, or similarity functions could yield different uncertainty trajectories. 

% What about the K values? 
Additionally, the semantic uncertainty measure we use is expensive relative to direct text to label prediction, specifically when sampling $K$ continuations per prefix (see Section~\ref{subsec:estimating_semantic_uncertainty}). The $K$ ablation suggests that this cost is reducible. Mean $F_{0.5}$ reaches 0.54 at $K=2$ and remains near its best values for small $K$, indicating that the coarse uncertainty measures may still be informative enough to predict TRPs locally. Regardless of smaller $K$ values, the per-prefix cost of sampling and embedding remains higher than single-pass prediction. Future work is required to determine whether semantic uncertainty can be deployed in a real-time system. 

% Q: What about the simple decision rule? 
Our analysis also relies on a deliberately simple decision mechanism. The fixed, causal rule used here (Section~\ref{subsec:decision_rule}) prioritizes interpretability over expressive capacity. The variance decomposition in Section~\ref{subsec:exp_SU} suggests this choice is consequential: the decision rule accounts for 75.4\% of $F_{0.5}$ variance across configurations, with smoothing alone explaining 52.6\%. More flexible models, which are capable of exploiting longer-range dependencies or finer-grained patterns in the uncertainty trajectory, could plausibly outperform a deterministic rule. The reported performance should therefore not be read as a performance upper bound on uncertainty-based TRP projection.

Finally, our evaluation metrics reflect a particular interactional cost structure. We prioritize precision-weighted measures because false positive TRP predictions are often more disruptive than missed opportunities to respond (Section~\ref{subsec:eval_metrics}). This assumption may not hold in all settings. More proactive or high-initiative systems may require different trade-offs between false positives and false negatives. Our metrics also assess local prediction quality, not downstream outcomes such as conversational fluidity or user experience. The results should therefore be interpreted as alignment with perceived TRPs under this cost structure, rather than as a complete measure of interactional success.

\section{Ethical Considerations}
\label{sec:ethical_considerations}

We consider the ethical implications of this work in terms of how turn prediction signals might be used in deployed systems. If such signals are treated as entitlements to speak rather than advisory cues, semantic uncertainty could lead to inappropriate interruptions or missed opportunities to respond. We therefore emphasize that these signals are model-mediated and should not be interpreted as evidence of user intent or readiness to yield the floor. While turn-taking is often described as universal, its realization varies across cultures and interactional settings, and systems that operationalize TRPs risk privileging particular norms if deployed without care. Additionally, the participant-judgment data used in this work were collected under ethical oversight, with informed consent, protections for participant anonymity, and approval from the relevant institutional review board.

\finalsection{
\section{Acknowledgments}
\label{sec:acknowledgements}

We thank Vasanth Sarathy for early discussions on structuring our experiments, Bilal Ahmed for refining key ideas and providing feedback, and Julia Mertens for thoughtful discussions that strengthened our arguments. We also acknowledge the Tufts University Department of Computer Science for institutional support, and Tufts Research Technology High Performance Computing for providing the computational resources used in this work. AI assistants were used in this work to improve language consistency; all scientific content and results are the authors' original work. 
}

% \nocite{Aho:72}
% \bibliography{acl-example}
\bibliography{refs}

\clearpage
\appendix

\section{Operationalizing TRPs: Stimulus-Specific Thresholding and Local Maximality}
\label{appsec:trp_label_construction}

% What are the ways in which we can operationalize our data? 
The nature of the listener-response data permits several operationalizations of binary TRP labels. A \emph{global threshold} labels an interval positive when its response proportion exceeds a fixed cutoff across all stimuli. A \emph{stimulus-specific threshold} labels an interval positive when its response proportion is elevated relative to the distribution for that stimulus. A \emph{local-peak criterion} labels an interval positive when its response proportion exceeds those of its adjacent intervals.

We use the stimulus-specific threshold described in Section~\ref{subsec:data}. This choice reflects two properties of listener responses. First, response onsets are temporally dispersed because listeners project TRPs and may begin responding before the stimulus turn ends \citep{sacks1974simplest,deRuiter2006ProjectingTheEnd,levinson2015timing}. High absolute agreement within any single interval is therefore unlikely. Second, stimuli differ in baseline responsiveness, making a fixed global threshold overly conservative for low-response stimuli and overly permissive for high-response stimuli. Stimulus-specific thresholding accounts for this variability.

\begin{figure}[h]
    \centering
    \includegraphics[width=\columnwidth]{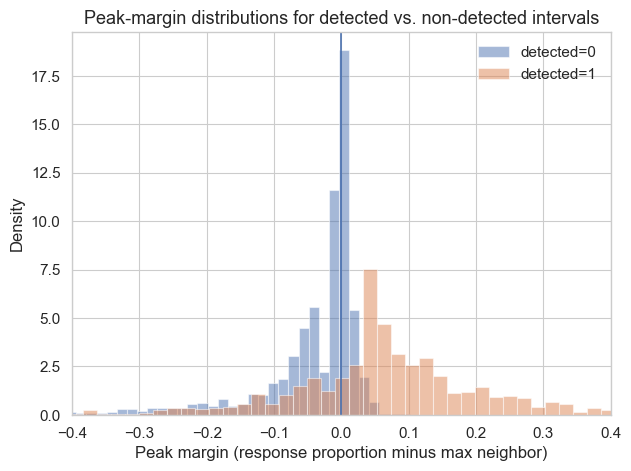}
    \caption{Distribution of peak margins for detected TRPs and non-TRP intervals. Peak margin is defined as the difference between an interval’s response proportion and the maximum of its immediate neighbors. Detected TRPs exhibit substantially larger positive margins, while non-TRP intervals are centered near zero, indicating that the labeling procedure selects positions with locally elevated listener response rates.}
    \label{fig:local_peak_salience}
\end{figure}

The stimulus-specific threshold identifies elevated response proportions but does not establish whether they form distinct peaks rather than broader regions of elevated responsiveness. For each interval, we therefore compute a \emph{local peak margin}: the difference between $\intervalProportion{i}{i+1}$ and the larger response proportion of its immediate neighbors, where defined. Positive margins indicate strict local maxima (see Figure~\ref{fig:local_peak_salience}). Among the 842 positive labels, 72\% occur at strict local maxima. This statistic is diagnostic, not definitional: local maximality is not part of the labeling rule. It shows that the threshold generally selects positions that are locally prominent in the response distribution.

\section{Model Selection and Experimental Infrastructure}
\label{appsec:models_and_infra}

% This appendix describes the experimental setup used throughout this work. Unless stated otherwise, all experiments follow the configuration outlined here. Any experiment-specific deviations from these defaults are detailed in the relevant appendices.

\subsection{Model Selection}

\begin{table}[t]
\centering
\small
\renewcommand{\arraystretch}{1.12}
\setlength{\tabcolsep}{4pt}
\begin{tabular}{@{}lccc@{}}
\toprule
\textbf{Model} & \textbf{Type} & \textbf{Params.} & \textbf{Precision} \\
\midrule
LLaMA-3.1-8B-Instruct
& Dense & 8B & q4f16 \\
LLaMA-3.1-70B-Instruct
& Dense & 70B & q4f16 \\
Mistral-7B-Instruct
& Dense & 7B & q4f16 \\
Mixtral-8$\times$7B-Instruct
& MoE & 46B & q4f16 \\
Qwen2.5-7B-Instruct
& Dense & 7B & q4f16 \\
Qwen2.5-72B-Instruct
& Dense & 72B & q4f16 \\
\bottomrule
\end{tabular}
\caption{
Instruction-tuned, open-weight LLMs used throughout this work. Dense models apply all parameters to each input, whereas Mixture-of-Experts (MoE) models route each input through a sparse subset of expert modules. Parameter counts for MoE models report total parameters across experts. q4f16 denotes 4-bit weight quantization with FP16 compute.
}
\label{tab:models_evaluated}
\end{table}

%Q: What models are used, and why these ones?
Throughout this work, we use a set of instruction-tuned, open-weight LLMs spanning multiple architectural families and parameter scales (see Table~\ref{tab:models_evaluated}). This selection tests whether observed behaviors vary systematically with model scale or architecture, rather than reflecting idiosyncrasies of a single model family. The set includes both dense and mixture-of-experts architectures, which allocate representational capacity differently and exhibit distinct scaling and generalization properties \citep{fedus2022switch}. Varying model scale probes whether limitations reflect insufficient capacity or the formulation of the TRP prediction task itself, since scaling can affect model capabilities unevenly across tasks \citep{wei2022emergent}. All models are instruction-tuned, which improves adherence to task specifications \citep{zhang2023instruction}.

\subsection{Experimental Infrastructure}

\begin{table*}[t]
\centering
\small
\renewcommand{\arraystretch}{1.14}
\setlength{\tabcolsep}{4.5pt}
\begin{tabular}{llrrrr}
\toprule
\textbf{Experiment} 
& \textbf{Stage} 
& \makecell{\textbf{\#}\\\textbf{Configs}} 
& \makecell{\textbf{Aggregate}\\\textbf{GPU-h}} 
& \makecell{\textbf{Aggregate}\\\textbf{CPU-h}} 
& \makecell{\textbf{Avg wall}\\\textbf{/ unit}} \\
\midrule
\multirow{4}{*}{\makecell[l]{Prompting\\baselines\\(\S\ref{subsec:exp_prompt_baselines})}}
 & Expert & 6 & 7.037 & 28.148 & 0.813 s/prefix \\
 & Participant & 6 & 10.932 & 43.728 & 1.26 s/prefix \\
 & Imagined & 6 & 11.755 & 47.020 & 1.36 s/prefix \\
 & Oracle & 6 & 20.078 & 80.312 & 2.32 s/prefix \\
\midrule
\multirow{2}{*}{\makecell[l]{SFT\\(\S\ref{subsec:exp_SFT})}}
 & Training & 3 & 0.591 & 2.366 & 11.83 min/model \\
 & Inference & 3 & 0.351 & 1.404 & 7.02 min/model \\
\midrule
\multirow{3}{*}{\makecell[l]{Semantic\\uncertainty\\(\S\ref{subsec:exp_SU})}}
 & Continuation sampling & 6 & 85.335 & 341.340 & 9.86 s/prefix \\
 & Embedding & 18 & 4.530 & 18.120 & 0.174 s/prefix \\
 & Decision function $\trpfunc$ & 2{,}304 & -- & -- & $<0.001$ s/prefix \\
\midrule
Total & -- & -- & 140.609 & 562.438 & -- \\
\bottomrule
\end{tabular}
\caption{
Compute summary for the main-text experiments. \# Configs reports the number of configurations or runs included in each stage. Aggregate GPU-h and CPU-h report total compute for that stage. Avg wall / unit reports average wall-clock time for the natural unit of the stage: per prefix for prompting and semantic-uncertainty stages, and per trained model for SFT stages. The semantic-uncertainty method is separated into continuation sampling, embedding, and decision-rule evaluation because only continuation sampling requires new LLM generations. The continuation-sampling row aggregates the six decoding configurations used in the fixed-$K{=}25$ sweep. Embedding and decision-rule configurations are computed from saved continuations, and decision-rule evaluation is CPU-only.
}
\label{tab:compute}
\end{table*}
% \begin{table*}[t]
% \centering
% \small
% \renewcommand{\arraystretch}{1.15}
% \setlength{\tabcolsep}{8pt}
% \begin{tabular}{l l c c c c}
% \toprule
% \textbf{Experiment} &
% \textbf{Stage} &
% \textbf{GPU-h} &
% \textbf{CPU-h} &
% \textbf{Peak GPU Mem (GB)} &
% \textbf{Host RAM (GB)} \\
% \midrule

% Prompting (\S\ref{subsec:exp_prompt_baselines})
%     & Inference  & 24.643 & 98.571  & 5.0--43.9 & 64--128 \\
% \midrule

% \multirow{2}{*}{SFT (\S\ref{subsec:exp_SFT})}
%     & Training   & 0.591  & 2.365  & --- & 128 \\
%     & Inference  & 0.351  & 1.405  & 7.7 & 128 \\
% \midrule

% \multirow{2}{*}{Semantic Uncertainty (\S\ref{subsec:exp_SU})}
%     & Cont.\ sampling & 85.335 & 341.340 & 7.7 & 64 \\
%     & Embedding       & 2.255  & 9.018   & --- & 64--128 \\
% \midrule

% \textbf{Total}
%     &  & \textbf{113.175} & \textbf{452.699} & 5.0--43.9 & 64--128 \\
% \bottomrule
% \end{tabular}

% \caption{
% Compute resources used across all experiments, aggregated by
% experiment and processing stage. Columns report GPU hours
% (GPU-h), CPU hours (CPU-h), peak GPU memory usage in gigabytes,
% and host memory requested for each run. All runs used a single
% GPU with four CPU cores. Experiments were executed on NVIDIA
% A100 (40GB and 80GB) and H200 GPUs. Stages with negligible
% compute cost are omitted.
% }
% \label{tab:compute}
% \end{table*}

% Q: How is the experimental infra. implemented? 
We implement efficient batched inference and consistent generation behavior across models using custom adaptations of the MLC-LLM framework together with the HuggingFace ecosystem \citep{wolf2019huggingface,mlc-llm}. Model weights are stored with 4-bit quantization and computation is performed in FP16, reducing memory usage and enabling large-scale evaluation on limited hardware. Parameter-efficient fine-tuning, which introduces low-rank trainable parameters to a frozen base model, is implemented using LoRA adapters \citep{hu2022lora}. Experiments run on NVIDIA A100 (40 GB and 80 GB) and H200 GPUs with a single accelerator per run and four CPU cores per GPU. Per-experiment aggregate compute usage is summarized in Table~\ref{tab:compute}. We do not include closed-source LLMs because they do not provide the transparency required for controlled scientific evaluation, including access to model weights, training data, and decoding settings \citep{liesenfeld2024rethinking}.

% Q: Why temperature and top-p as decoding parameters?
Across experiments, we use two decoding parameters, temperature and $top$-$p$, to control LLM generation stochasticity. Temperature rescales the shape of the full token distribution, while $top$-$p$ governs its effective support by restricting sampling to the smallest set of tokens whose cumulative probability exceeds a threshold \citep{holtzman2019curious}. Other decoding parameters (e.g., $top$-$k$ truncation or repetition penalties) use default values to limit interacting degrees of freedom and to preserve a consistent sampling regime. Specific settings are reported in each experiment's appendix.

% ---
\section{Prompt-Based TRP Prediction Details}
\label{appsec:prompt_trp_full_results}

In Section~\ref{subsec:exp_prompt_baselines}, we evaluate all LLMs (see Table~\ref{tab:models_evaluated}) under the four prompting conditions using a fixed decoding configuration (temperature = 0.4, $top$-$p$ = 1.0; \citealp{holtzman2019curious}). This allows us to balance diversity and determinism while treating generation stochasticity as a controlled source of variation rather than an object of analysis. We generate a single prediction per interval; sampling multiple generations would substantially increase computational cost across the full model–condition grid. As a result, reported metrics reflect a single sampled generation, and systematic analysis of decoding variability is left to future work.

For each prefix, prompts instruct models to produce structured JSON outputs containing a binary TRP decision, a confidence score, and a brief justification. Evaluation uses only the binary decision; confidence scores and justifications are recorded but excluded from reported metrics. Outputs are parsed using a deterministic post-processing routine, achieving a 99.5\% success rate over 124,680 generations. The remaining failures primarily reflect minor deviations from the expected schema and are excluded prior to evaluation; given their low frequency, they are unlikely to affect comparative results. Appendix~\ref{app:prompts} shows the summarized prompt templates for each condition.

\section{Supervised Fine-Tuning Details}
\label{appsec:SFT_details}

In Section~\ref{subsec:exp_SFT}, we fine-tune LLaMA-3.1-8B-Instruct using LoRA adapters, rather than the strongest prompt-based baseline model (LLaMA-3.1-70B-Instruct; see Section~\ref{subsec:exp_prompt_baselines}).  Prompt-based results indicate that larger models do not necessarily improve performance on the within-turn TRP prediction task. Fine-tuning a larger model could in principle yield different results, as supervised adaptation and prompting exhibit distinct failure modes; we leave this for future work.

% Q: What loss is used, and how does it relate to the task definition? 
We fine-tune using a causal language modeling objective with a completion-only loss, computing gradients only over the assistant completion corresponding to the binary TRP decision. Restricting gradients to the completion avoids updating the model on prompt or context tokens that are not part of the decision target. We do not introduce class-weighted or precision-aware losses, despite evaluating with $F_{0.5}$. Our aim is not to optimize downstream metrics but to test whether supervision alone improves TRP prediction under a standard SFT formulation. Incorporating task-specific loss shaping would confound this diagnostic.

% Q: How is training and inference actually performed? 
Training uses the AdamW optimizer with a base learning rate of $2\times10^{-5}$, cosine scheduling with a warmup ratio of 0.03, per-device batch size of 4, and gradient accumulation over 8 steps. Models are trained for up to 30 epochs with early stopping based on the completion-only evaluation loss, and the best checkpoint is selected for evaluation. At inference time, we generate TRP predictions using the same system prompt used during training, along with the same decoding parameters (temperature = 0.4, $top$-$p$ = 1.0) as used for prompt-based inference (see Appendix~\ref{appsec:prompt_trp_full_results}).

%----- UPDATED 

\section{Semantic Uncertainty Based TRP Prediction Details}
\label{appsec:SU_exp_detais}

\begin{table*}[t]
\centering
\small
\renewcommand{\arraystretch}{1.12}
\setlength{\tabcolsep}{3.5pt}
\begin{tabularx}{\textwidth}{@{}
    p{0.20\textwidth}
    p{0.23\textwidth}
    X
    c
@{}}
\toprule
\textbf{Stage} & \textbf{Hyperparameter} & \textbf{Values} & \textbf{\#} \\
\midrule

\shortstack[l]{Uncertainty\\estimation\\(Algorithm~\ref{alg:semantic_uncertainty})}
& Generation model
& LLaMA-3.1-8B-Instruct (q4f16)
& 1 \\

& Continuations per prefix
& $K=25$
& 1 \\

& Decoding setting $(T,\mathrm{top}\text{-}p)$
& $(0.50,0.80)$; $(0.80,0.90)$; $(0.85,0.95)$; $(0.90,0.90)$; $(0.90,1.00)$; $(1.10,1.00)$
& 6 \\

& Embedding model
& SFR-2R; all-mpnet-base-v2; E5-Mistral-7B-Instruct
& 3 \\

& SNNE scale ($\tau$)
& $\{0.1,1,10,100\}$
& 4 \\

\midrule

\shortstack[l]{Decision\\function\\(Algorithm~\ref{alg:trp_decision})}
& Smoothing
&
\begin{tabular}[t]{@{}l@{}}
EMA: $\alpha \in \{0.3,0.5,0.7\}$; \\
Boxcar: window $\in \{3,5,7\}$; \\
Gaussian: $\sigma=1.5$, window $\in \{5,10\}$
\end{tabular}
& 8 \\

& Rise/drop windows
& Rise window $\in \{5,10\}$ crossed with drop window $\in \{5,10\}$
& 4 \\

& Fixed detector settings
& Causal operation; adaptive thresholding; threshold $=1.5$; minimum local points $=3$; derivative-based rise/drop scoring; median centering; MAD-to-$\sigma$ factor $=1.4826$; minimum scale $\varepsilon=10^{-8}$
& 1 \\

\bottomrule
\end{tabularx}
\caption{
Main fixed-$K$ robustness sweep, organized as a two-phase pipeline. In Phase~1, uncertainty estimation crosses 6 decoding settings, 3 embedding models, and 4 SNNE $\tau$ values, yielding 72 uncertainty-signal configurations. In Phase~2, the decision function converts each trajectory into TRP predictions using 32 detector configurations, formed by crossing 8 causal smoothing variants with 4 adaptive rise/drop window settings. In total, the sweep evaluates $72 \times 32 = 2{,}304$ scored configurations.
}
\label{tab:semantic_uncertainty_sweep}
\end{table*}

% ----

% This appendix provides implementation details and
% diagnostic analyses for the semantic-uncertainty experiment in Section~\ref{subsec:exp_SU}. Shared infrastructure and compute reporting follow Appendix~\ref{appsec:models_and_infra}.

\subsection{Experimental Setup}
\label{subsec:SU_exp_setup_details}

% Q: What are we doing / how many params? 
We evaluate the semantic uncertainty method (see Section~\ref{subsec:projecting_trps_via_semantic_uncertainty}) by sweeping three groups of parameters: continuation-sampling settings, semantic-uncertainty settings, and decision-rule settings. Table~\ref{tab:semantic_uncertainty_sweep} summarizes the configuration space.

% Q: What sampling model do we use and why? 
We use LLaMA-3.1-8B-Instruct as the continuation model. This model provides a practical trade-off between instruction-following ability and computational cost. Empirical runtimes indicate that
LLaMA-3.1-70B-Instruct is between 6$\times$ and 20$\times$ slower than its smaller variant. We therefore use the 8B model throughout and vary continuation diversity through six temperature and top-$p$ settings.

% Q: How do we embed and what models? 
We embed continuations using three different embedding models. We include \texttt{SFR-Embedding-2\_R} as a general-purpose embedding model for semantic similarity and retrieval, and \texttt{all-mpnet-base-v2} as a smaller sentence-transformer baseline \citep{reimers2019sentence}. We also include \texttt{intfloat/e5-mistral-7b-instruct} as a larger embedding model, allowing us to test whether the uncertainty signal depends on embedding-model scale \citep{wang2024improving}. 

For each model, we compute SNNE from pairwise cosine similarities between continuation embeddings. While \citet{nguyen-etal-2025-beyond} report ROUGE-L as the strongest similarity function for their task, we use embedding cosine similarity because our continuations are short ($\approx$5–10 tokens) and often differ in surface form. In this setting, lexical-overlap measures may be sensitive to small wording differences rather than semantic similarity \citep{callisonburch2006reevaluating}.

\subsection{Semantic Uncertainty Robustness}
\label{subsec:SU_robustness}

% What do we do in this analysis? 
To assess whether the semantic-uncertainty signal is sensitive to continuation sampling, we summarize its behavior across the six decoding configurations (see Table~\ref{tab:semantic_uncertainty_sweep}). We hold $K=25$ and $\tau=0.1$ fixed to isolate the effect of temperature and top-$p$.

First, we summarize the uncertainty trajectory for each decoding setting. For each embedding model, we compute SNNE for every prefix, then measure the mean and standard deviation across prefixes (see Table~\ref{tab:decoding_config_summary_avg}). As decoding becomes more stochastic, mean SNNE becomes less negative, from $-11.74$ to $-10.61$, indicating broader continuation sets. The standard deviation across prefixes decreases from $0.56$ to $0.21$, indicating compressed prefix-level contrast.

\begin{table}[t]
\centering
\small
\setlength{\tabcolsep}{3pt}
\renewcommand{\arraystretch}{1.05}
\begin{tabular}{@{}ccccc@{}}
\toprule
\multicolumn{2}{c}{\textbf{Decoding}} & \multicolumn{3}{c}{\textbf{Metric, mean $\pm$ SD}} \\
\cmidrule(lr){1-2}\cmidrule(l){3-5}
$T$ & $p$ & SNNE & Pref. Std & Cos. \\
\midrule
.50 & .80 & $-11.74 \pm .35$ & $.56 \pm .09$ & $.72 \pm .17$ \\
.80 & .90 & $-10.99 \pm .51$ & $.41 \pm .00$ & $.62 \pm .22$ \\
.85 & .95 & $-10.99 \pm .53$ & $.39 \pm .01$ & $.62 \pm .23$ \\
.90 & .90 & $-10.87 \pm .51$ & $.36 \pm .02$ & $.60 \pm .23$ \\
.90 & 1.00 & $-10.77 \pm .50$ & $.30 \pm .04$ & $.58 \pm .25$ \\
1.10 & 1.00 & $-10.61 \pm .46$ & $.21 \pm .06$ & $.54 \pm .27$ \\
\bottomrule
\end{tabular}
\caption{
Decoding diagnostic for fixed $K{=}25$ and $\tau{=}0.1$. SNNE is the mean uncertainty value across prefixes. Pref. Std is the standard deviation of SNNE across prefixes. Cos. is the average cosine similarity between distinct continuations sampled for the same prefix. Values are mean $\pm$ SD across the three embedding models.
}
\label{tab:decoding_config_summary_avg}
\end{table}

Second, we measure semantic diversity within each prefix under the same decoding setting. For each prefix, we compute the average cosine similarity between all distinct pairs of continuations sampled for that prefix, then average across prefixes and embedding models. This continuation similarity decreases from $0.72$ to $0.54$ as decoding becomes more stochastic, indicating that higher stochasticity produces more semantically diverse continuations for the same prefix.

Finally, we test whether the same prefix remains stable across decoding settings. For each prefix and decoding setting, we average the embeddings of its sampled continuations. We then compare these mean embeddings for the same prefix across every pair of the six decoding configurations. Mean cosine similarity exceeds $0.94$, indicating that changing temperature and top-$p$ increases continuation diversity without moving the same prefix to a substantially different semantic region.

These diagnostics show that changing temperature and top-$p$ makes the continuations broader and more diverse, but the same prefix still points to a similar semantic region across decoding settings. The uncertainty trajectory is therefore not just an artifact of one sampling setup; it reflects information tied to the prefix itself.

\subsection{SNNE Similarity Scaling and Discriminative Capacity}
\label{subsec:tau_analysis}

% Q: What are we analyzing?
Our method predicts within-turn TRPs from local changes in the semantic-uncertainty trajectory. The SNNE similarity scaling parameter $\tau$ controls the numerical scale of this trajectory. Smaller values weigh similar continuation pairs highly, while larger values distribute weight more evenly and compress differences among SNNE values.

% Q: How did we do the analysis? 
To quantify this effect, we compute a reference contrast for each $\tau$ using two synthetic $K=25$ similarity matrices. One represents complete agreement, with $\prefixSimMatrix{i}{u,v}=1$ for all $u,v$; the other is a positive-similarity dispersion reference, with $\prefixSimMatrix{i}{u,u}=1$ and $\prefixSimMatrix{i}_{u,v}=0$ for $u\neq v$. This contrast does not define the full SNNE range, since cosine similarities can be negative, but it provides a fixed calibration of how much numerical contrast SNNE expresses as $\tau$ varies. We compare it to the observed standard deviation of prefix-level SNNE values from the fixed-$K$ sweep, averaged over embedding models and decoding configurations.

Table~\ref{tab:tau_sensitivity} shows that the reference contrast shrinks rapidly as $\tau$ increases. This confirms that large $\tau$ values compress SNNE scale. At $\tau=100$, the prefix-level standard deviation exceeds the reference contrast, meaning that this calibration no longer reflects the full empirical range of the signal. The main sweep shows, however, that this compression has limited downstream effect. The higher-level implication is that $\tau$ primarily affects the scale of the uncertainty trajectory, while TRP prediction depends on how local changes in that trajectory are interpreted by the decision rule.

\begin{table}[H]
\centering
\small
\renewcommand{\arraystretch}{1.15}
\setlength{\tabcolsep}{5pt}
\begin{tabular}{r r r r}
\toprule
$\tau$ &
\textbf{Ref.\ Contrast} &
\textbf{Prefix Std} &
\textbf{Std / Contrast} \\
\midrule
0.1 & 3.218 & 0.371 & 11.5\% \\
1   & 0.934 & 0.078 & 8.4\% \\
10  & 0.096 & 0.022 & 23.3\% \\
100 & 0.010 & 0.020 & 206.8\% \\
\bottomrule
\end{tabular}
\caption{
Effect of the similarity scaling parameter $\tau$ on SNNE
discriminative capacity for $K=25$ continuations. Reference contrast is the SNNE difference between identical
($\prefixSimMatrix{i}_{u,v}=1$ for all $u,v$) and orthogonal
($\prefixSimMatrix{i}_{u,u}=1$, $\prefixSimMatrix{i}_{u,v}=0$ for
$u\neq v$) similarity configurations. Prefix Std denotes the observed standard deviation of SNNE across prefixes.
}
\label{tab:tau_sensitivity}
\end{table}

\subsection{In-Domain Semantic-Similarity}
\label{appsubsec:embedding_validity}

Section~\ref{subsec:estimating_semantic_uncertainty} treats embedding-space dispersion as a model-mediated proxy for semantic variation, without assuming independence from lexical form. Prior work supports this interpretation: Sentence-BERT is designed so that cosine similarity reflects semantic similarity \citep{reimers2019sentence}, E5-Mistral is evaluated on semantic-similarity and retrieval tasks \citep{wang2024improving}, and Semantic Textual Similarity benchmarks compare model similarities with human judgments \citep{cer2017semeval}.

However, these evaluations largely concern complete sentences or passages, whereas Algorithm~\ref{alg:semantic_uncertainty} embeds LLM-generated continuations of approximately 5--10 tokens conditioned on prefixes transcribed from spontaneous speech. We therefore conduct a targeted in-domain diagnostic to test whether the embedding space captures semantic similarity in this setting.

We sample candidate continuations across all six decoding settings (see Table~\ref{tab:semantic_uncertainty_sweep}). For each continuation, the comparison reported here uses a triplet comprising the original, a similar-length paraphrase preserving its meaning while changing its wording, and a similar-length continuation from another stimulus with a different meaning. We retain 78 cases after automatic checks for comparable length, distinct wording, and the absence of prefix copying, followed by manual verification of the intended semantic relationships. For example, ``was a really tough experience'' is paired with the paraphrase ``was incredibly challenging and difficult'' and the unrelated continuation ``from the main hallway suddenly.'' We embed the continuations without their prefixes using \texttt{SFR-Embedding-2\_R}, one of the embedding models evaluated in the main experiment (see Appendix~\ref{subsec:SU_exp_setup_details}).

In 77 of 78 triplets, the original is more similar to its paraphrase than to the unrelated continuation (98.7\%; 95\% paired-bootstrap CI: 96.2--100\%). Mean cosine similarity is 0.861 for original--paraphrase pairs and 0.654 for original--unrelated pairs, with a mean paired difference of 0.207 (95\% CI: 0.190--0.223). Thus, for the short continuations used here, the embedding space captures semantic similarity across lexical variation, providing support for its use as a model-mediated proxy.

\clearpage

\makeatletter
\newcommand{\appletterlabel}[1]{%
  \phantomsection%
  \protected@edef\@currentlabel{\Alph{section}}% what \ref{#1} will print
  \label{#1}%
}
\makeatother

\setcounter{figure}{0}
\renewcommand{\thefigure}{\arabic{figure}}

\refstepcounter{section}
\begin{figure*}[!t]
\captionsetup{justification=raggedright,singlelinecheck=false}

\caption*{\large\bfseries ~\Alph{section} Prompt Templates}
\appletterlabel{app:prompts}

\centering
\begin{tcolorbox}[
  skin=standard,
  colback=gray!25,
  colframe=gray!80,
  coltitle=black,
  colbacktitle=gray!45,
  fonttitle=\bfseries,
  title={Prompt 1: Expert (Theory-Guided TRP Judgment)},
  boxrule=0.7pt,
  arc=2mm,
  left=7pt, right=7pt,
  top=6pt, bottom=6pt,
  toptitle=3pt, bottomtitle=3pt,
]
\small

\vspace{0.5em}
\noindent\textbf{[ROLE]}\vspace{0.2em}\par\noindent
You are a Conversation Analysis (CA) expert evaluating, incrementally after each prefix, whether a Transition Relevance Place (TRP) occurs \textbf{immediately after the final word}. 
A TRP is an opportunity (not an obligation) for speaker transition or for the current speaker to begin a new Turn Construction Unit (TCU).\par

\vspace{0.5em}
\noindent\textbf{[SOURCES]}\vspace{0.2em}\par\noindent
\textbf{Note: For clarity of presentation, the academic sources originally included in the prompt have been omitted here.}

\vspace{0.5em}
\noindent\textbf{[TASK]}\vspace{0.2em}\par\noindent
Given a sequence of words spoken by a single speaker (\texttt{<PREFIX>}),
decide whether it ends at a point where a listener could appropriately
begin speaking or where the current speaker could transition to a new
Turn Construction Unit.\par

\vspace{0.5em}
\noindent\textbf{[DECISION RUBRIC]}\vspace{0.2em}\par\noindent
A TRP is an \emph{opportunity}, not an obligation, for speaker transition.
\begin{itemize}[leftmargin=1.6em, itemsep=0pt, topsep=2pt, parsep=0pt, partopsep=0pt]
  \item \textbf{Output 1 (TRP)} when the utterance is syntactically and pragmatically complete (e.g., an independent clause or completed social action).
  \item \textbf{Output 0 (no TRP)} when the utterance projects continuation (e.g., open coordination or subordination, discourse markers, or continuation punctuation).
  \item If cues conflict, prefer \textbf{0} unless completion is decisively clear.
\end{itemize}

\vspace{0.5em}
\noindent\textbf{[OUTPUT FORMAT]}\vspace{0.2em}\par\noindent
\begin{verbatim}
{"does_trp_occur":0 or 1,"confidence":0.0-1.0,
 "justification":"brief explanation (<=1000 chars)"}
\end{verbatim}
The output must end with the token \texttt{<END>}.
The confidence and justification fields are recorded for analysis
but are not used for post-processing or decision thresholding.\par

\vspace{0.5em}
\noindent\textbf{[EXAMPLE (SCHEMATIC)]}\vspace{0.2em}\par\noindent
Illustrative examples use \textbf{fabricated text}:
\begin{itemize}[leftmargin=1.6em, itemsep=0pt, topsep=2pt, parsep=0pt, partopsep=0pt]
  \setlength{\itemsep}{1pt}
  \item \texttt{"I think we're done."} $\rightarrow$ \texttt{does\_trp\_occur = 1}
  \item \texttt{"I was thinking that"} $\rightarrow$ \texttt{does\_trp\_occur = 0}
\end{itemize}

\end{tcolorbox}
\phantomcaption
\label{fig:expert_prompt}
\end{figure*}

\begin{figure*}[!t]
\centering
\begin{tcolorbox}[
  skin=standard,
  colback=gray!25,
  colframe=gray!80,
  coltitle=black,
  colbacktitle=gray!45,
  fonttitle=\bfseries,
  title={Prompt 2: Participant (Intuitive TRP Judgment)},
  boxrule=0.7pt,
  arc=2mm,
  left=7pt, right=7pt,
  top=6pt, bottom=6pt,
  toptitle=3pt, bottomtitle=3pt,
]
\small

\vspace{0.5em}
\noindent\textbf{[ROLE]}\vspace{0.2em}\par\noindent
You are taking the role of a participant in a conversation study.
You imagine listening to a speaker and deciding whether you could naturally
produce a brief encouraging response (e.g., ``yeah'', ``mmhmm'') at the end
of what they just said.\par

\vspace{0.5em}
\noindent\textbf{[TASK]}\vspace{0.2em}\par\noindent
Given a single line of speech spoken by one speaker (\texttt{<PREFIX>}),
decide whether it ends at a point where you could \textbf{reasonably give a short
listener response}. Always make a decision for the final word of the line.\par

\vspace{0.5em}
\noindent\textbf{[DECISION RUBRIC]}\vspace{0.2em}\par\noindent
A TRP is an \emph{opportunity}, not an obligation, for speaker transition.
\begin{itemize}[leftmargin=1.6em, itemsep=0pt, topsep=2pt, parsep=0pt, partopsep=0pt]
  \item \textbf{Output 1 (TRP)} when the utterance is syntactically and pragmatically complete (e.g., an independent clause or completed social action).
  \item \textbf{Output 0 (no TRP)} when the utterance projects continuation (e.g., open coordination or subordination, discourse markers, or continuation punctuation).
  \item If cues conflict, prefer \textbf{0} unless completion is decisively clear.
\end{itemize}

\vspace{0.5em}
\noindent\textbf{[OUTPUT FORMAT]}\vspace{0.2em}\par\noindent
\begin{verbatim}
{"does_trp_occur":0 or 1,"confidence":0.0-1.0,
 "justification":"brief explanation (<=1000 chars)"}
\end{verbatim}
The output must end with the token \texttt{<END>}.
The confidence and justification fields are recorded for analysis
but are not used for post-processing or decision thresholding.\par

\vspace{0.5em}
\noindent\textbf{[EXAMPLE (SCHEMATIC)]}\vspace{0.2em}\par\noindent
Illustrative examples use \textbf{fabricated text}:
\begin{itemize}[leftmargin=1.6em, itemsep=0pt, topsep=2pt, parsep=0pt, partopsep=0pt]
  \item \texttt{"I think we're done."} $\rightarrow$ \texttt{does\_trp\_occur = 1}
  \item \texttt{"I was thinking that"} $\rightarrow$ \texttt{does\_trp\_occur = 0}
\end{itemize}

\end{tcolorbox}
\phantomcaption
\label{fig:participant_prompt}
\end{figure*}

\begin{figure*}[!t]
\centering
\begin{tcolorbox}[
  skin=standard,
  colback=gray!25,
  colframe=gray!80,
  coltitle=black,
  colbacktitle=gray!45,
  fonttitle=\bfseries,
  title={Prompt 3: Participant (Imagined Future Continuation)},
  boxrule=0.7pt,
  arc=2mm,
  left=7pt, right=7pt,
  top=6pt, bottom=6pt,
  toptitle=3pt, bottomtitle=3pt,
]
\small

\vspace{0.3em}
\noindent\textbf{[ROLE]} \vspace{0.2em}\par\noindent
You are listening to a single speaker in a natural conversation and must judge
whether the end of the speaker’s turn-so-far is a natural point for a brief listener response.
\par

\vspace{0.3em}
\noindent\textbf{[TASK]} \vspace{0.2em}\par\noindent
First, imagine how the speaker is likely to continue next by writing a short
continuation in the speaker’s voice (1--2 clauses, $\approx$10 words). Then,
decide whether the end of the given line is a point where a brief listener
response would naturally fit.

\vspace{0.3em}
\noindent\textbf{[DECISION GUIDELINES]}\vspace{0.2em}\par\noindent
\begin{itemize}[leftmargin=1.6em, itemsep=0pt, topsep=2pt, parsep=0pt, partopsep=0pt]
  \item \textbf{Output 1 (TRP)} if the turn-so-far sounds complete and the continuation would be optional.
  \item \textbf{Output 0 (no TRP)} if the turn-so-far leads directly into the provided continuation.
  \item Short items (e.g., speaker acknowledgments) may be complete on their own.
\end{itemize}

\vspace{0.3em}
\noindent\textbf{[OUTPUT FORMAT]}\vspace{0.2em}\par\noindent
\begin{verbatim}
{"does_trp_occur":0 or 1,
 "projected_upcoming_turns":"imagined speaker continuation",
 "confidence":0.0-1.0,
 "justification":"brief explanation (<=1000 chars)"}
\end{verbatim}
The output must end with the token \texttt{<END>}.
The projected continuation is used only as part of the judgment and is
not evaluated against ground truth.\par

\vspace{0.3em}
\noindent\textbf{[EXAMPLE (SCHEMATIC)]}\vspace{0.2em}\par\noindent
Illustrative examples use \textbf{fabricated text}:
\begin{itemize}[leftmargin=1.6em, itemsep=0pt, topsep=2pt, parsep=0pt, partopsep=0pt]
  \item \texttt{"I think we're done."} $\rightarrow$
        continuation: \texttt{"We can leave whenever you're ready."},
        \texttt{does\_trp\_occur = 1}
  \item \texttt{"I was thinking that"} $\rightarrow$
        continuation: \texttt{"we should try something else."},
        \texttt{does\_trp\_occur = 0}
\end{itemize}
\end{tcolorbox}
\phantomcaption
\label{fig:imagined_future_prompt}
\end{figure*}

\begin{figure*}[!t]
\centering
\begin{tcolorbox}[
  skin=standard,
  colback=gray!25,
  colframe=gray!80,
  coltitle=black,
  colbacktitle=gray!45,
  fonttitle=\bfseries,
  title={Prompt 4: Participant (Oracle Future Continuation)},
  boxrule=0.7pt,
  arc=2mm,
  left=7pt, right=7pt,
  top=6pt, bottom=6pt,
  toptitle=3pt, bottomtitle=3pt,
]
\small

\vspace{0.3em}
\noindent\textbf{[ROLE]}\vspace{0.2em}\par\noindent
You are listening to a single speaker in a natural conversation and must judge
whether the end of the speaker’s turn-so-far is a natural point for a brief listener response.
\par

\vspace{0.3em}
\noindent\textbf{[TASK]}\vspace{0.2em}\par\noindent
You are given:
\begin{enumerate}[leftmargin=1.6em, itemsep=0pt, topsep=2pt]
  \item a line of speech spoken by one speaker (\texttt{<PREFIX>}), and
  \item the \textbf{actual upcoming continuation} of the same speaker.
\end{enumerate}
Use the provided continuation \textbf{exactly as written} to decide whether
the end of the given line is a point where a brief listener response would
naturally fit.\par

\vspace{0.3em}
\noindent\textbf{[DECISION GUIDELINES]}\vspace{0.2em}\par\noindent
\begin{itemize}[leftmargin=1.6em, itemsep=0pt, topsep=2pt, parsep=0pt, partopsep=0pt]
  \item \textbf{Output 1 (TRP)} if the turn-so-far sounds complete and the continuation would be optional.
  \item \textbf{Output 0 (no TRP)} if the turn-so-far leads directly into the provided continuation.
  \item Short items (e.g., speaker acknowledgments) may be complete on their own.
\end{itemize}

\vspace{0.3em}
\noindent\textbf{[OUTPUT FORMAT]}\vspace{0.2em}\par\noindent
\begin{verbatim}
{"does_trp_occur":0 or 1,
 "projected_upcoming_turns":"verbatim provided continuation",
 "confidence":0.0-1.0,
 "justification":"brief explanation (<=1000 chars)"}
\end{verbatim}
The output must end with the token \texttt{<END>}.
The provided continuation must be copied verbatim and is not generated by the model.
The confidence and justification fields are recorded for analysis but are not used
for post-processing or decision thresholding.\par

\vspace{0.3em}
\noindent\textbf{[EXAMPLE (SCHEMATIC)]}\vspace{0.2em}\par\noindent
Illustrative examples use \textbf{fabricated text}:
\begin{itemize}[leftmargin=1.6em, itemsep=0pt, topsep=2pt, parsep=0pt, partopsep=0pt]
  \item Turn-so-far: \texttt{"I think we're done."} \\
        Continuation: \texttt{"We can head out whenever you're ready."} \\
        $\rightarrow$ \texttt{does\_trp\_occur = 1}
  \item Turn-so-far: \texttt{"I was thinking that"} \\
        Continuation: \texttt{"we might need to try a different approach."} \\
        $\rightarrow$ \texttt{does\_trp\_occur = 0}
\end{itemize}
\end{tcolorbox}
\phantomcaption
\label{fig:oracle_future_prompt}
\end{figure*}

\end{document}